\documentclass[final]{cvpr}

\usepackage{times}
\usepackage{epsfig}
\usepackage{graphicx}
\usepackage{amsmath}
\usepackage{amssymb}
\usepackage{bm}
\usepackage{booktabs}
\usepackage{subfigure}
\usepackage{multirow}
\usepackage[dvipsnames]{xcolor}
\usepackage{float}
\usepackage{caption}

\newcommand{\norm}[1]{\left\lVert#1\right\rVert}

\usepackage[pagebackref=true,breaklinks=true,colorlinks,bookmarks=false]{hyperref}

\def\cvprPaperID{1223} % *** Enter the CVPR Paper ID here
\def\confYear{CVPR 2021}
\begin{document}

%%%%%%%%% TITLE
\title{Unsupervised Disentanglement of Linear-Encoded Facial Semantics}

\author{Yutong Zheng, Yu-Kai Huang, Ran Tao, Zhiqiang Shen and Marios Savvides\\
Carnegie Mellon University\\
{\tt\small \{yutongzh,yukaih2,rant,zhiqians,marioss\}@andrew.cmu.edu}
% \and
% Yu-Kai Huang\\

% \and
% Ran Tao\\

% \and
% Zhiqiang Shen\\

% \and
% Marios Savvides\\
}

% \author{\IEEEauthorblockN{Yutong Zheng, Yu-Kai Huang, Ran Tao, Zhiqiang Shen and Marios Savvides}
% \IEEEauthorblockA{Department of Electrical and Computer Engineering,
% Carnegie Mellon University\\
% Email: \IEEEauthorrefmark{1}author.one@add.on.net}}

% \maketitle
\twocolumn[{%
\maketitle
\begin{center}
    \centering
    \includegraphics[width=2\columnwidth]{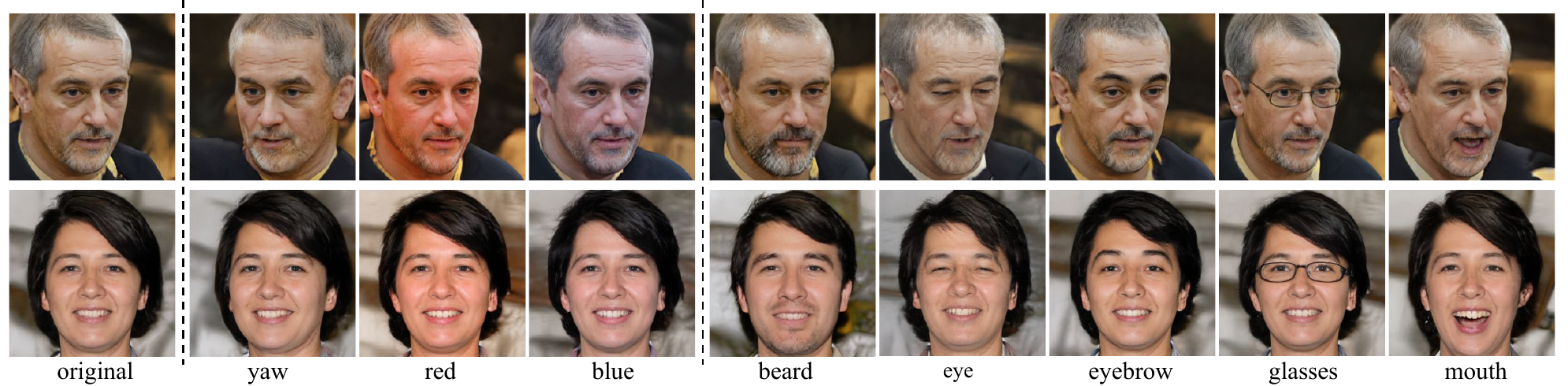}
    \captionof{figure}{Facial manipulation result by sampling along latent representations in StyleGAN, learned by our unsupervised framework. \textbf{First column}: Original generations. \textbf{Column 2-4}: manipulation on univariate environmental semantics. \textbf{Column 5-9}: manipulation on localized facial semantics. }\label{fig:projected_sample}
\end{center}%
}]

%%%%%%%%% ABSTRACT
\begin{abstract}
   We propose a method to disentangle linear-encoded facial semantics from StyleGAN without external supervision. The method derives from linear regression and sparse representation learning concepts to make the disentangled latent representations easily interpreted as well. We start by coupling StyleGAN with a stabilized 3D deformable facial reconstruction method to decompose single-view GAN generations into multiple semantics. Latent representations are then extracted to capture interpretable facial semantics. In this work, we make it possible to get rid of labels for disentangling meaningful facial semantics. Also, we demonstrate that the guided extrapolation along the disentangled representations can help with data augmentation, which sheds light on handling unbalanced data. Finally, we provide an analysis of our learned localized facial representations and illustrate that the semantic information is encoded, which surprisingly complies with human intuition. The overall unsupervised design brings more flexibility to representation learning in the wild. 
\end{abstract}

%%%%%%%%% BODY TEXT

\section{Introduction}
In recent years, Generative Adversarial Networks (GANs) \cite{goodfellow2014generative} have been a great success in synthesizing photo-realistic images given a set of latent codes. Despite the rapid boost in image quality, the interpretability of the generation process has become another major area of research. In general, interpretability requires latent codes to encode disentangled semantic information of the image. Further, ideally, well-disentangled semantics are supposed to be factorized to practically interpretable components and each component should be linear-encoded in the latent space as representation \cite{higgins2016beta,desjardins2012disentangling,chen2018isolating,kim2018disentangling,rezende2014stochastic,eastwood2018framework}. 
\thispagestyle{empty}

StyleGAN \cite{karras2019style} proposes a new architecture by bringing an intermediate latent space to provide support for disentanglement property for face generation. Consequently, facial semantics are linear-encoded as latent representations. Based on StyleGAN, recent works show that sampling along the linear-encoded representation vector in latent space will change the associated facial semantics accordingly \cite{shen2020interpreting}, which makes it possible to manipulate the face generations to meet a target requirement. However, in current frameworks, mapping a particular facial semantics to a latent representation vector relies on training offline classifiers with manually labeled datasets. Thus they require artificially defined semantics and provide the associated labels for all facial images. The disadvantages for training with labeled facial semantics include: first, they demand extra effort on human annotations for each new attributes proposed; second, each semantics is defined artificially, and the scope of semantics is limited to the linear combination of such definitions; and third, by only training on each labeled semantics independently, we are unable to give any insights on the connections among different semantics. 

In this work, we explore unsupervised methods to minimize the demand for human annotation. We propose a novel unsupervised framework to disentangle and manipulate facial semantics under the StyleGAN environment, while still maintain the interpretability for semantics (Fig.~\ref{fig:projected_sample}) as in labeled datasets. 
\begin{itemize}
    \item We motivate \textit{decorrelation regularization} on StyleGAN to further enhance disentanglement for the latent representation. 
    \item We introduce \textit{mutual reconstruction} to stabilize training of an unsupervised 3D deformable face reconstruction method, such that it serves as an initial facial semantic extractor. 
    \item For univariate semantics, e.g., yaw angle, we present a \textit{linear regression method} to capture their perturbations from latent space. Given the manipulation vector, we further demonstrate the success of yaw manipulation on data augmentation to trivialize the unbalanced pose problem within the unsupervised paradigm. 
    \item For pixel-level semantics, e.g., shape and texture, we propose a \textit{localized representation learning algorithm} to capture sparse semantic perturbations from latent space. The associated analysis evinces the effectiveness of our method to provide interpretability without external supervision. 
\end{itemize}
All methods proposed are purely based on a label-free training strategy. Only StyleGAN is trained with an in-the-wild face dataset. Therefore, we reduce a significant amount of human involvement in facial representation learning. Furthermore, with zero labels, our framework provides an unconstrained environment for the disentanglement algorithms to explore and shed light on how interpretable representations are learned in cutting-edge neural network models. 

%-------------------------------------------------------------------------

\section{Related Works}

\subsection{Exploring latent space representations in GANs}
GAN \cite{goodfellow2014generative} is widely explored on applications with the photo-realistic images \cite{gulrajani2017improved, martin2017wasserstein, miyato2018spectral, zhang2019self}. The understanding of how GANs learn to construct the latent space with the semantics information defined in the real visual world draws attention recently \cite{shen2020interpreting, shen2020interfacegan}.

Prior exploration of the latent space of GANs focuses on smoothly varying the output image from one synthesis to another with the interpolation in the latent space without considering the underlying semantics \cite{shao2018riemannian, laine2018feature}. 
A simultaneous optimization strategy on the generator and latent code is developed to learn a better constructed latent space\cite{bojanowski2017optimizing}. Other works with recurrent neural networks also explore the latent code space with the semantics information. One of which, \cite{jahanian2019steerability} interprets the steer-ability from the perspective of camera motion and image color tone. Another \cite{yang2019semantic} uses the hierarchical semantics to understand the deep generative representations for scene synthesis. There are also explorations using facial attributes for face synthesis, e.g. InterFaceGAN \cite{shen2020interpreting}, which provided a detailed analysis of the semantics encoded in the latent space. They take into consideration both single and disentangled multi-semantics and, by a fixed pretrained GAN, it learns to explore semantics by relating the latent space with labeled semantic attributes on the synthesized images. DiscoFaceGAN \cite{deng2020disentangled} uses the pretrained 3DMM \cite{blanz1999morphable} parameters to provide guidance for interpretable face manipulation. Overall, The major commonsense for the learning of latent space semantic representation is that it is supervised and only in that way, the semantics are interpretable and controllable. 

The editing of synthesized faces with GANs is another active research area. Face editing is generally conducted with semantic information such as facial attributes. With facial attributes act together, face images can be therefore constructed. Thus, face editing requires the GANs to have the ability to edit the disentangled information. In other words, the editing should change the target attribute but keep other information ideally unchanged. To achieve this effect, methods are mainly focused on three aspects: the comprehensive design of loss functions \cite{chen2016infogan, odena2017conditional,tran2017disentangled}, the involvement of additional attribute features \cite{shen2018facefeat, bao2018towards,lample2017fader,yin2017towards} and the architecture designs \cite{shen2018faceid, donahue2017semantically, liu2019stgan, zhu2017unpaired, choi2018stargan}. However, these works either discard the latent code, resulting in the inability to continuously interpolate certain semantics, or fail to provide the same synthesis quality compared with state-of-the-art GANs \cite{karras2019style, karras2020analyzing}. 

\subsection{Unsupervised disentanglement method}
Generally speaking, all the vanilla GANs are unsupervised representation learning methods because they only require training on unlabeled datasets. Variational auto-encoders (VAEs) \cite{kingma2013auto}, on the other hand, achieves a similar goal with self-reconstruction loss. Recent efforts have been made to make VAEs generate images as good as GANs \cite{vahdat2020nvae, van2017neural}. Unsupervised disentanglement methods both exists in GANs \cite{chen2016infogan, lin2019infogan} and VAEs \cite{higgins2016beta, kim2018disentangling}. However, most of these unsupervised disentangled methods only work under simple datasets and are considered not disentangle realistic information \cite{locatello2019challenging}. According to \cite{locatello2019challenging}, one should always introduce either supervision or inductive biases to the disentanglement method to achieve meaningful representations. 

Examples of inductive bias rise from the symmetry of natural objects and the 3D graphical information. Recent works successfully reconstruct the face images by carefully remodeling the graphics of camera principal \cite{wu2020unsupervised}. This work makes it possible to decompose the facial images into environmental semantics and other facial semantics. However, a major drawback of this work is that it is unable to generate realistic faces and perform pixel-level face editing on it. 

\section{Method}
The goal of our facial representation disentanglement method is to capture linear-encoded facial semantics. With a given collection of coarsely aligned faces, a Generative Adversarial Network (GAN) is trained to mimic the overall distribution of the data. In our case, to better learn linear-encoded facial semantics, we re-implement and train StyleGAN-2 \cite{karras2020analyzing} as it is well known for the ability to disentangle representations in latent space. Further, we investigate the latent space trained by StyleGAN and improve its capability to disentangle by adding a decorrelation regularization (Sec.~\ref{method:reg}). 
After training a StyleGAN model, we use the faces it generates as training data and trains a 3D deformable face reconstruction method modified from \cite{wu2020unsupervised}. A mutual reconstruction strategy (Sec.~\ref{method:3dface}) stabilizes the training significantly.  
Then, we keep a record of the latent code from StyleGAN and apply linear regression to disentangle the target semantics in the latent space (Sec.~\ref{method:linear_regression}). Meanwhile, taking the reconstruction of the yaw angle as an example, we manipulate the latent representation as a data augmentation for training (Sec.~\ref{method:dataaug}). 
Finally, we describe the localized representation learning method to disentangle canonical semantics in Sec.~\ref{method:canonical}. 

\begin{figure}[t]
\begin{center}
    \subfigure[original training protocol fails.]{
        \includegraphics[trim={0 9.35cm 0 0.05cm},clip, width=0.8\columnwidth]{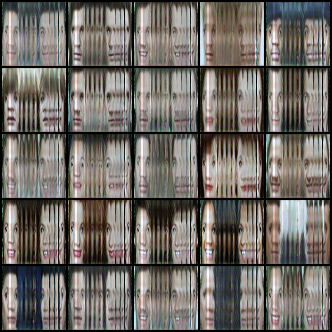}\label{fig:corrupted_albedo}}\\[-1ex]
    \subfigure[mutual reconstruction strategy stabilizes training.]{
        \includegraphics[trim={0 9.35cm 0 0.05cm},clip, width=0.8\columnwidth]{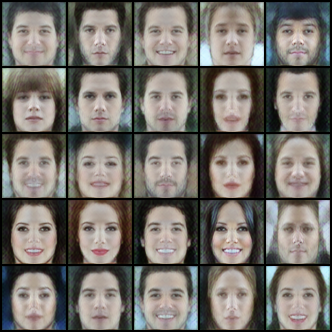}\label{fig:correct_albedo}}
\end{center}
\vspace{-0.7cm}
\caption{The canonical albedo map of samples in CASIA WebFace after training for 60k iterations with batch size of 64.}\label{fig:mutual_reconstruction}
\vspace{-0.5cm}
\end{figure}

\subsection{Decorrelating latent code in StyleGAN}
\label{method:reg}
In StyleGAN design, a latent code $\mathbf{z}\in\mathcal{Z}^{d\times1}$ is randomly generated from a Gaussian distribution. Then, a mapping network takes in $\mathbf{z}$ and output a latent code $\mathbf{w}\in\mathcal{W}^{d\times1}$. Space $\mathcal{W}$ is proven to facilitate the learning of more disentangled representations. In our study, we find that we can further enhance the disentangled representation by decorrelating latent codes in $\mathcal{W}$. Intuitively, a more decorrelated latent space enforces more independent dimensions to encode information, therefore encourages disentanglement in representations. 
In order to maximize the utilization of all dimensions in $\mathcal{W}$, we want to make all Pearson correlation coefficient $\rho_{ij}$ to zero and variance of all dimensions $\mathrm{Var}[w_i]$ the same value, where $i,j$ stand for the subscript of dimensions in $\mathcal{W}$ space. Therefore, we introduce decorrelation regularization via a loss function:
\begin{align}
\begin{split}
    \mathcal{L}_{decorr} = &-\sum_{i\ne j}\left(\log{\left(1-|\rho_{ij}|\right)}\right)\\
    &+ \sum_{i}\left(\mathrm{Var}[w_i]-\sum_{j}\left(\mathrm{Var}[w_j]\right)\right)^2.
\end{split}
\label{eqn:decorr}
\end{align}
Here, $\rho_{ij}$ and $\mathrm{Var}[w_i]$ are all estimated by sampling $\mathbf{w}$ from the mapping network $\mathcal{F}(\mathbf{z})$, given $\mathbf{z}\sim \mathcal{N}(\mathbf{0},\,\mathbf{I})$. 

The overall objective for GAN with decorrelation regularization follows: 
\begin{align*}
    \min_{\mathcal{F},\mathcal{G}}{\max_\mathcal{D}{\mathcal{L}_{GAN}}+\mathcal{L}_{decorr}},
\end{align*}
where $\mathcal{G}$ and $\mathcal{D}$ stand for the generator and discriminator for GAN, respectively. Here the mapping network $\mathcal{F}$ is the only one to update with the new loss, $\mathcal{L}_{decorr}$.

\subsection{Stabilized training for 3D face reconstruction}
\label{method:3dface}
The unsupervised 3D deformable face reconstruction method \cite{wu2020unsupervised} takes a roughly aligned face image and decomposes the faces into multiple semantics, i.e. view, lighting, albedo, and depth ($\mathbf{y}_{v}, \mathbf{y}_{l}, \mathbf{y}_{a}$ and $\mathbf{y}_{d}$, respectively). During training, it uses these decomposed semantics to reconstruct the original input image $I$ with the reconstruction loss: 
\begin{align*}
    \mathcal{L}_{recon} &= \mathcal{C}(I)^\top |I-\hat{I}|,\\
    \text{where} \ \hat{I} &= \mathcal{R}(\mathbf{y}_{v}, \mathbf{y}_{l}, \mathbf{y}_{a}, \mathbf{y}_{d}),
\end{align*}
where $\mathcal{C}$ and $\mathcal{R}$ stand for a confidence network and a 3D face renderer. We use this method to pre-decompose some external facial semantics, i.e. pose and lighting, from StyleGAN generations.

\begin{figure}[t]
\begin{center}
        \includegraphics[width=\columnwidth]{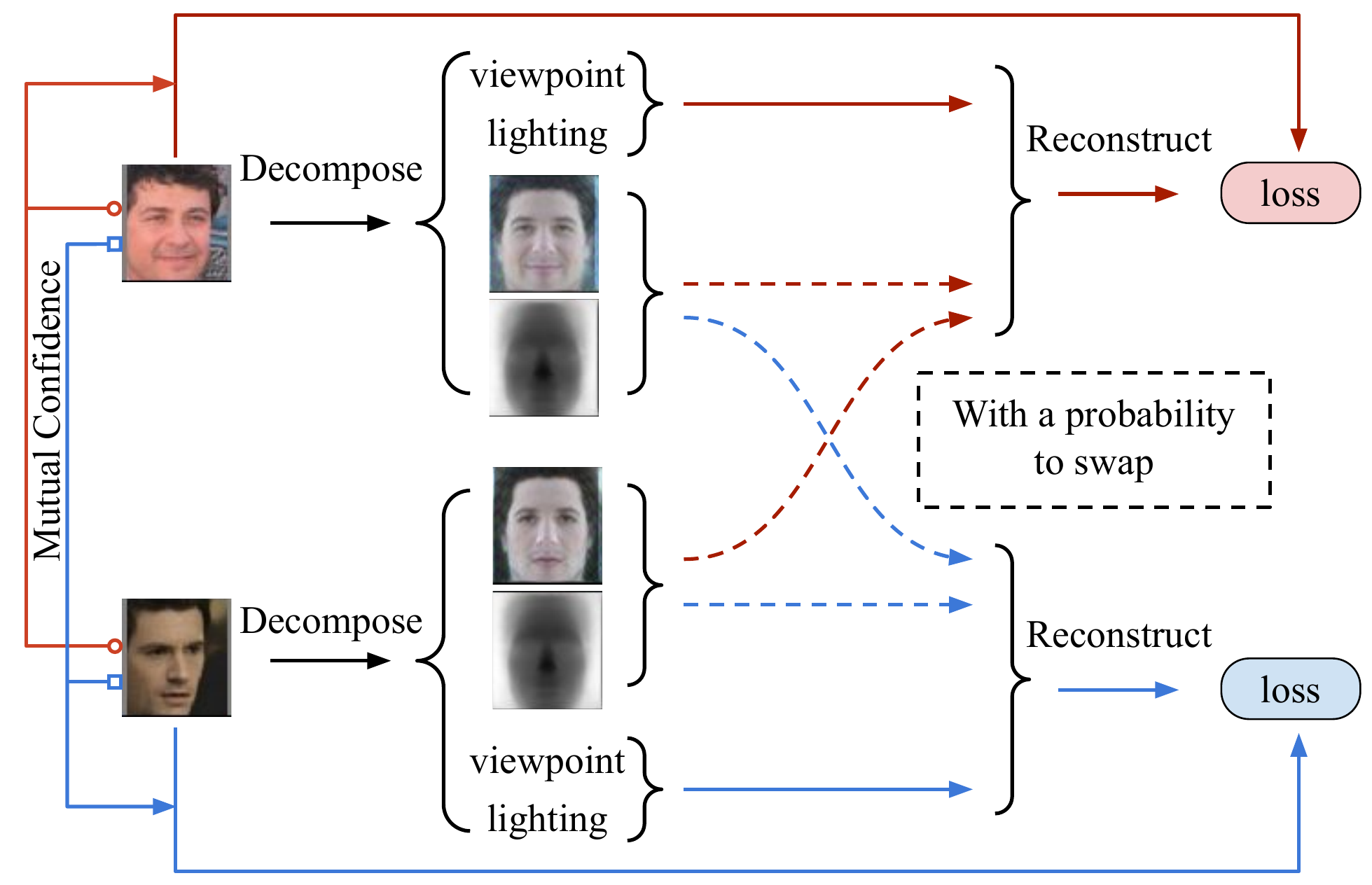}
\end{center}
\vspace{-0.5cm}
\caption{The implementation of mutual reconstruction during training. }\label{fig:mutual_reconstruction_figure}
\vspace{-0.5cm}
\end{figure}

However, we find that the 3D face reconstruction algorithm struggles to estimate the pose of profile or near-profile faces. This finding coincides with the failure case analysis in the original paper \cite{wu2020unsupervised}. To make things worse, if the dataset contains a decent number of profile and near-profile faces (e.g. CASIA WebFace), the 3D reconstruction fails to learn physically sounded semantics (Shown in Fig.~\ref{fig:corrupted_albedo}) and collapses into a sub-optimal state. That is, the algorithm tries to use extreme values to estimate the texture and shape of each face \textit{independently}, which deviate far away from the actual texture and shape of the face. 

To deal with the problem, we develop a mutual reconstruction strategy as illustrated in Fig.~\ref{fig:mutual_reconstruction_figure}. To start with, we know that in general, human faces are very similar in shape and texture. Therefore, each face image should still be able to reconstruct itself comparatively if we swap its shape and texture with another face. Eventually, the strategy prevents the model from using extreme values to fit individual reconstruction, and the model learns to reconstruct faces with a minimum variance of the shape and texture among all samples, even for profile faces. Following this idea, during training, we swap the albedo and depth map between two images with a probability $\epsilon$ to perform the reconstruction with the alternative loss:
\begin{align*}
    \Tilde{\mathcal{L}}_{recon} &= \mathcal{\Tilde{C}}(I, I')^\top |I-\hat{I}'|,\\
    \text{where} \ \hat{I}' &= \mathcal{R}(\mathbf{y}_{v}, \mathbf{y}_{l}, \mathbf{y}'_{a}, \mathbf{y}'_{d}),
\end{align*}
where $\mathcal{\Tilde{C}}$ is a mutual confidence network; and everything with the prime notation originates from another image $I'$. The overall loss to reconstruct each image becomes:
\begin{align*}
    (1-\epsilon)\mathcal{L}_{recon} + \epsilon\Tilde{\mathcal{L}}_{recon}.
\end{align*}
As a result, the shape and texture of faces with deviated facial semantics can be robustly estimated. 

Moreover, since images are now reconstructed with two images, the confidence map in the original work should be yielded by these two images accordingly. We simply concatenate the two images channel-wise as input to the confidence network, where the top image provides environmental semantics and the bottom image provides texture and shape information. 

\subsection{Disentangle semantics with linear regression}
\label{method:linear_regression}
With the 3D face reconstruction algorithm, face images generated by StyleGAN are decomposed to pose, lighting, depth, and albedo. Remember, the ultimate goal of disentangling semantics is to find a vector $\mathbf{v} \in \mathcal{W}$ in StyleGAN, such that it only takes control of the target semantics. 
%We divide all the available semantics into two groups: independent univariate semantics and pixel-level canonical semantics. 

\textbf{Semantic gradient estimation:} Now consider a semantics $y$ of a generated face image $\mathcal{G}(\mathbf{w})$ that can be measured by a function $f(\cdot)$. The linear approximation of the gradient $\bm{\nabla}_y$ with respect to the latent code $\mathbf{w}$ satisfies:
\begin{align}
    f(\mathcal{G}(\mathbf{w}_1)) \approx f(\mathcal{G}(\mathbf{w}_0)) + \bm{\nabla}_y(\mathbf{w}_0)^\top (\mathbf{w}_1 - \mathbf{w}_0).\nonumber
\end{align}
% Independent univariate semantics such as facial pose (yaw, pitch, and roll angles), light direction, and strength are theoretically independent of each other. Thus we will treat them independently for the following algorithm. 
Note that, generally, the gradient at location $\mathbf{w}_0$, $\bm{\nabla}_y(\mathbf{w}_0)$, is a function of latent code $\mathbf{w}_0$. However, with StyleGAN, it is observed that many semantics can be linear-encoded in the disentangled latent space $\mathcal{W}$ \cite{karras2019style,shen2020interpreting}. We take advantage of this merit from StyleGAN and assume all the semantics can be linear-encoded. In other words, the gradient is no now independent of the input latent code $\mathbf{w}_0$. We get:
\begin{align}
    f(\mathcal{G}(\mathbf{w}_1)) \approx f(\mathcal{G}(\mathbf{w}_0)) + \bm{\nabla}_y^\top (\mathbf{w}_1 - \mathbf{w}_0),\nonumber
\end{align}
or to be simplified as:
\begin{align}
    \Delta y \approx \bm{\nabla}_y^\top \Delta\mathbf{w},\nonumber
\end{align}
where $\Delta y=f(\mathcal{G}(\mathbf{w}_1))-f(\mathcal{G}(\mathbf{w}_0))$ and $\Delta\mathbf{w}=\mathbf{w}_1 - \mathbf{w}_0$. 

\textbf{Semantic linear regression:} Now it is obvious that in the ideal case, the target vector $\mathbf{v}=\bm{\nabla}_y$. While in real world scenario, the gradient $\bm{\nabla}_y$ is hard to estimate directly because back-propagation only captures local gradient, making it less robust to noises. Therefore we propose a linear regression model to capture global linearity for gradient estimation. We randomly sample $N$ pairs of $(\mathbf{w}_1, \mathbf{w}_0)$, generate images with StyleGAN and estimate its semantics. Finally, all samples of differences are concatenated, denoted as $\Delta Y\in\mathbb{R}^{N\times1}$ for semantics and $\Delta\mathbf{W}\in\mathbb{R}^{N\times d}$ for latent codes. The objective is to minimize:
\begin{align}
    \min_{\mathbf{v}}~~ \norm{\Delta Y - \Delta\mathbf{W}\mathbf{v}}_2^2.
    \label{eqn:lregreg}
\end{align}
We have a closed-form solution when $N>d$:
\begin{align}
    \mathbf{v} = (\Delta\mathbf{W}^\top\Delta\mathbf{W})^{-1}\Delta\mathbf{W}^\top\Delta Y.
    \label{eqn:gradestimation}
\end{align}

\subsection{Image manipulation for data augmentation:}
\label{method:dataaug}
One useful application for guided image manipulation is to perform data augmentation. Data augmentation has proven to be efficient when dealing with unbalanced data during training. One related problem within our unsupervised framework is the inaccurate estimation of extreme yaw angle, which is also mentioned in \cite{wu2020unsupervised}. This problem worsens when dealing with generations from CASIA StyleGAN since it contains a large amount but a small portion of profile faces (unbalanced yaw distribution). 

In our experiment setting, we propose a data augmentation strategy base on self-supervised facial image manipulation. The end goal is to help the 3D face reconstruction network estimate the extreme yaw angle accurately. With the linear regression method mentioned in Sec.~\ref{method:linear_regression}, we can learn manipulation vectors $\mathbf{v}$ for univariate semantics including the yaw angle, denoted as $\mathbf{v}_{yaw}$. Recall that by extrapolating along $\mathbf{v}$ beyond its standard deviation, preferably, we can get samples with more extreme values for the associated semantics. Particularly, we can generate images with an extreme yaw angle to neutralize the unbalanced yaw distribution and train the 3D face reconstruction algorithm. Therefore, by seeing more profile faces deliberately, the system can better estimate extreme yaw angles. 

We perform the data augmentation strategy alongside with the training of 3D face reconstruction. To be specific, we estimate $\mathbf{v}$ and update with a historical moving average (momentum$=0.995$) every 10 iterations. Then, augmentation is achieved by extrapolating a latent vector $\mathbf{w}_i^{(s)}$ via:
\begin{align}
    \mathbf{w}_i^{(s)} = \mathbf{w}_i - \mathbf{w}_i^\top \mathbf{v}\mathbf{v} + s\cdot\sigma_{\mathbf{w}}\mathbf{v},
    \label{eqn:manipulate}
\end{align}
where $\mathbf{w}_i$ is a random sample drawn from $\mathcal{F}(\mathbf{z})$, $s$ is the scaling factor for the interpolation/extrapolation along the unit length manipulation vector $\mathbf{v}$. And $\sigma_{\mathbf{w}}$ is the standard deviation for $\mathbf{w}^\top \mathbf{v}$. In this case $\mathbf{v}=\mathbf{v}_{yaw}$. Finally, the 3D face reconstruction method is trained with the augmented generations $\mathcal{G}(\mathbf{w}_i^{(s)})$. 

\begin{table}[t]
\caption{Results on Decorrelation Regularization in StyleGAN}
\vspace{-0.3cm}
\label{exp:tab_decorr}
\centering
\begin{tabular}{@{}ccc|cc@{}}
\toprule
\multirow{2}{*}{\textbf{\begin{tabular}[c]{@{}c@{}}Decorrelated\\ latent codes\end{tabular}}} & \multicolumn{2}{c}{\textbf{CASIA} $(96\times112)$} & \multicolumn{2}{c}{\textbf{FFHQ} $(256^2)$} \\ \cmidrule(l){2-5} 
                & FID              & PPL             & FID             & PPL             \\ \midrule
No              & 11.85            & 15.26           & 25.49           &  34.24          \\ \midrule
Yes             & \textbf{9.96}    & \textbf{12.30}  & \textbf{15.57}  & \textbf{20.23}  \\ \bottomrule
\end{tabular}
\vspace{-0.3cm}
\end{table}

\subsection{Localized representation learning}
\label{method:canonical}
In the case where $f(\cdot)$ returns canonical outputs, i.e. depth and albedo maps, the outputs consist of pixels in spatial dimensions and the pixel values are highly correlated as the latent code changes. However, every pixel-level gradient estimation, i.e. $\mathbf{v}$, from Eqn.~\ref{eqn:gradestimation} is independently calculated and is thus extremely redundant. To deal with this problem, we reformulate our goal for canonical semantics; that is, to find the manipulation vectors $\hat{\mathbf{v}}$ that capture interpretable \textit{combinations} of pixel value variations. We start by defining a Jacobian matrix $\mathbf{J}_{\mathbf{v}}\in\mathbb{R}^{S\times d}$, which is the concatenation of all canonical pixel-level $\mathbf{v}$. Here $S$ stands for the overall number of spatial and RGB dimensions of a given depth and albedo map.

\begin{figure}[t]
\centering
\includegraphics[trim={0.05cm 0 0 0},clip,width=\columnwidth]{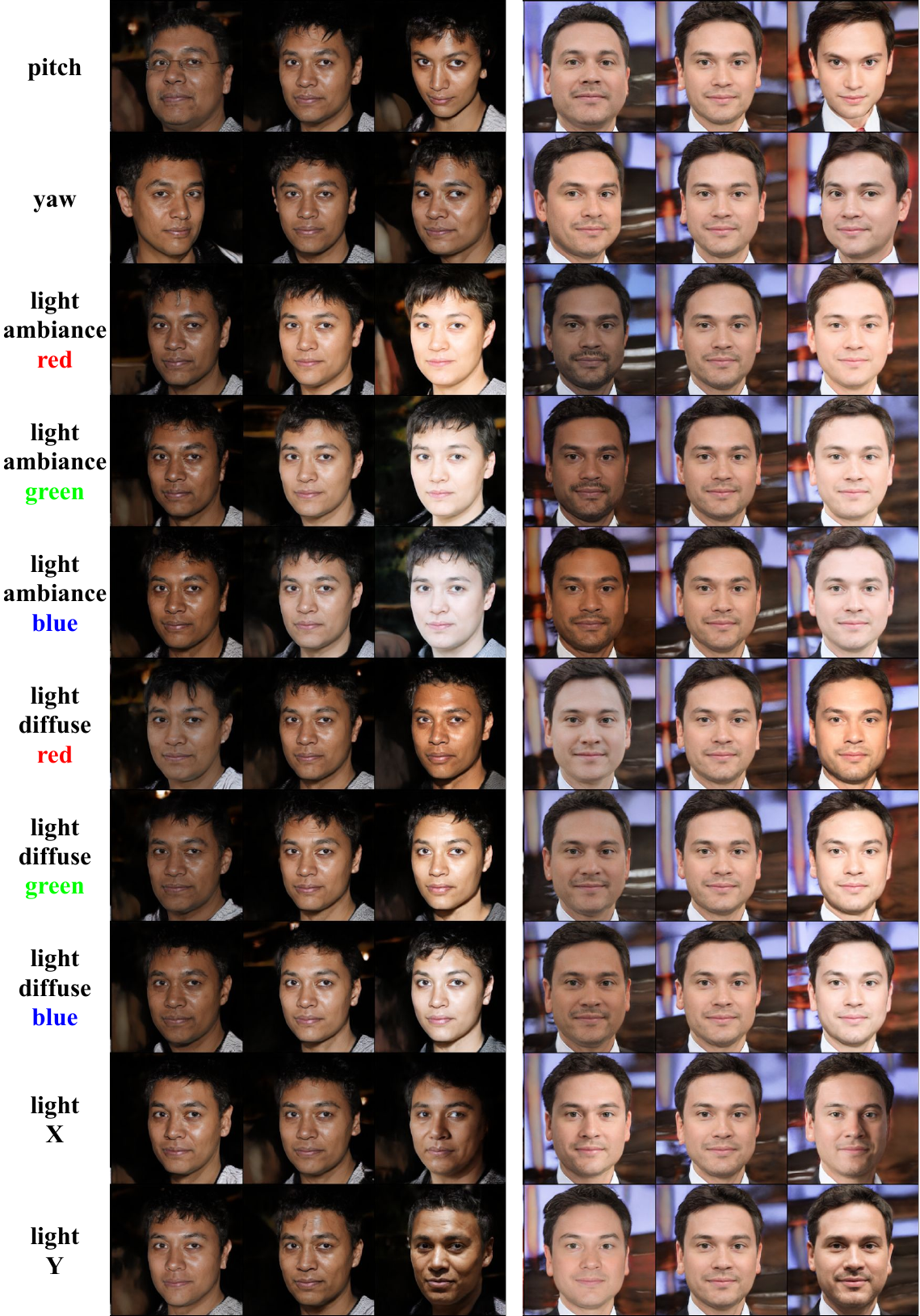}
\vspace{-0.7cm}
\caption{Two examples of the manipulation result on univariate semantics in StyleGAN trained with FFHQ.}\label{fig:univariate_semantics}
\vspace{-0.5cm}
\end{figure}

One trivial definition of $\hat{\mathbf{v}}$ is that it maximizes $\norm{\mathbf{J}_{\mathbf{v}}^*\hat{\mathbf{v}}}_2^2$. However, we need to keep in mind that, ideally we expect a disentangled representation to manipulate \textit{interpretable} facial semantics. That is to say, interpolation along $\hat{\mathbf{v}}$ should result in significant but \textit{localized} (i.e. sparse) change across the image domain, e.g. some $\hat{\mathbf{v}}$ only control eye variations while some only control mouth variations, etc.. However, $\norm{\mathbf{J}_{\mathbf{v}}^*\hat{\mathbf{v}}}_2^2$ captures the \textit{global} pixel value perturbation. Thus, we propose the \textbf{localized canonical representation learning} by solving: 
\begin{align}
\begin{split}
    \min_{\mathbf{U}, \hat{\mathbf{V}}}~~ &\ \norm{\mathbf{J}_{\mathbf{v}}^*-\mathbf{U}\hat{\mathbf{V}}^\top}_F^2+\alpha\norm{\mathbf{U}}_1+\beta\sum_{i\ne j}{(\hat{\mathbf{v}}_i^\top\hat{\mathbf{v}}_j)^2}\\
    \text{s.t.} &\ \norm{\hat{\mathbf{v}}_p}_2=1,
\end{split}
\label{eqn:sparse_pca}
\end{align}
where $p,i,j\in\left\{1, \cdots, P\right\}$ and $P$ is the number of components to learn. Each column in $\mathbf{U}=[\mathbf{u}_1, \cdots, \mathbf{u}_P]\in\mathbb{R}^{S\times P}$ is a sparse component of the canonical albedo and depth perturbation, and $\hat{\mathbf{V}}=[\hat{\mathbf{v}}_1, \cdots, \hat{\mathbf{v}}_P]\in\mathbb{R}^{d\times P}$ consists of the associated latent representation in $\mathcal{W}$ space. $\alpha$ and $\beta$ are tuning parameters to control the trade-off among the reconstruction accuracy of $\mathbf{J}_{\mathbf{v}}^*$, sparsity of perturbation in semantics and orthogonality of associated latent representations.

\section{Experiments}

The datasets used for training StyleGAN is CASIA WebFace \cite{yi2014learning} or Flickr-Face-HQ (FFHQ). We choose the two datasets since they represent very different types of face images, CASIA WebFace is at low resolution and high pose variation, while FFHQ is at high resolution, high environmental variations but mostly frontal faces. We demonstrate the effectiveness of our unsupervised disentanglement method in both cases. For CASIA WebFace, we roughly align and crop the faces to $112\times96$ such that the faces are all in the center of the image with minimum roll angle variations. The images are resized to $128\times128$ before being fed into StyleGAN for training. For FFHQ, we resize the original images to $256\times256$ for training StyleGAN. All StyleGAN generations are resized to $64\times64$ to decompose into semantics via the 3D reconstruction algorithm. 

\begin{table}[t]
\caption{Correlation coefficients between learned latent representations and ground truth facial semantics.}
\vspace{-0.6cm}
\label{tab:uni_coeff}
\begin{center}
\begin{tabular}{@{}cll@{}}
\toprule
\multirow{2}{*}{\textbf{latent representations}} & \multicolumn{2}{c}{\textbf{facial semantics}}       \\ \cmidrule(l){2-3} 
                                                 & \multicolumn{1}{c}{pitch} & \multicolumn{1}{c}{yaw} \\ \midrule
pitch                                            & \textbf{0.7647}           & 0.0146                  \\ \midrule
yaw                                              & 0.0390                    & \textbf{0.9458}         \\ \midrule
ambiance lighting red                            & 0.0195                    & 0.1001                  \\ \midrule
ambiance lighting green                          & 0.0414                    & 0.0913                  \\ \midrule
ambiance lighting blue                           & 0.0134                    & 0.1175                  \\ \midrule
diffuse light red                                & 0.0528                    & 0.0402                  \\ \midrule
diffuse light green                              & 0.1051                    & 0.1243                  \\ \midrule
diffuse light blue                               & 0.0823                    & 0.1250                  \\ \midrule
light direction X                                & 0.0338                    & 0.2316                  \\ \midrule
light direction Y                                & 0.2166                    & 0.0057                  \\ \bottomrule
\end{tabular}
\end{center}
\vspace{-0.5cm}
\end{table}

\subsection{Effectiveness of decorrelation regularization}
We follow the implementation and keep all parameters the same as in StyleGAN-2 \cite{karras2020analyzing} and train until the model sees a total of 25 million images. The decorrelation regularization applies its update purely on the mapping network of StyleGAN. We test the performance of the model in terms of FID \cite{heusel2017gans} and PPL \cite{karras2019style} metrics.

The results shown in Table \ref{exp:tab_decorr} indicate that by decorrelating the latent codes in $\mathcal{W}$, the generated images become more realistic, meanwhile, the latent representations are more disentangled. 

\subsection{Stabilized 3D face reconstruction}
CASIA WebFace \cite{yi2014learning} is a more challenging dataset for the original 3D face reconstruction algorithm \cite{wu2020unsupervised} since it contains faces with extreme pose variations. We illustrate the difficulty by training the algorithm with its default parameter setting. Note that the default setting successfully trained with the FFHQ dataset in our experiments and CelebA \cite{liu2015faceattributes} dataset as reported in their original work. 

We stabilize the training process by applying the mutual reconstruction strategy and set the swap probability $\epsilon=0.5$ for CASIA WebFace. All faces are decomposed into reasonable semantics with a stabilized estimation of depth and albedo (examples are shown in Fig.~\ref{fig:mutual_reconstruction}). We also set the swap probability $\epsilon=0.1$ for FFHQ training for a more smoothed estimation of the depth and albedo. 

\begin{figure}[t]
\begin{center}
\subfigure[The correlation of estimated yaw angle vs the ground truth yaw angle.]{
    \includegraphics[trim={0 0.4cm 0 0.5cm},clip,width=\columnwidth]{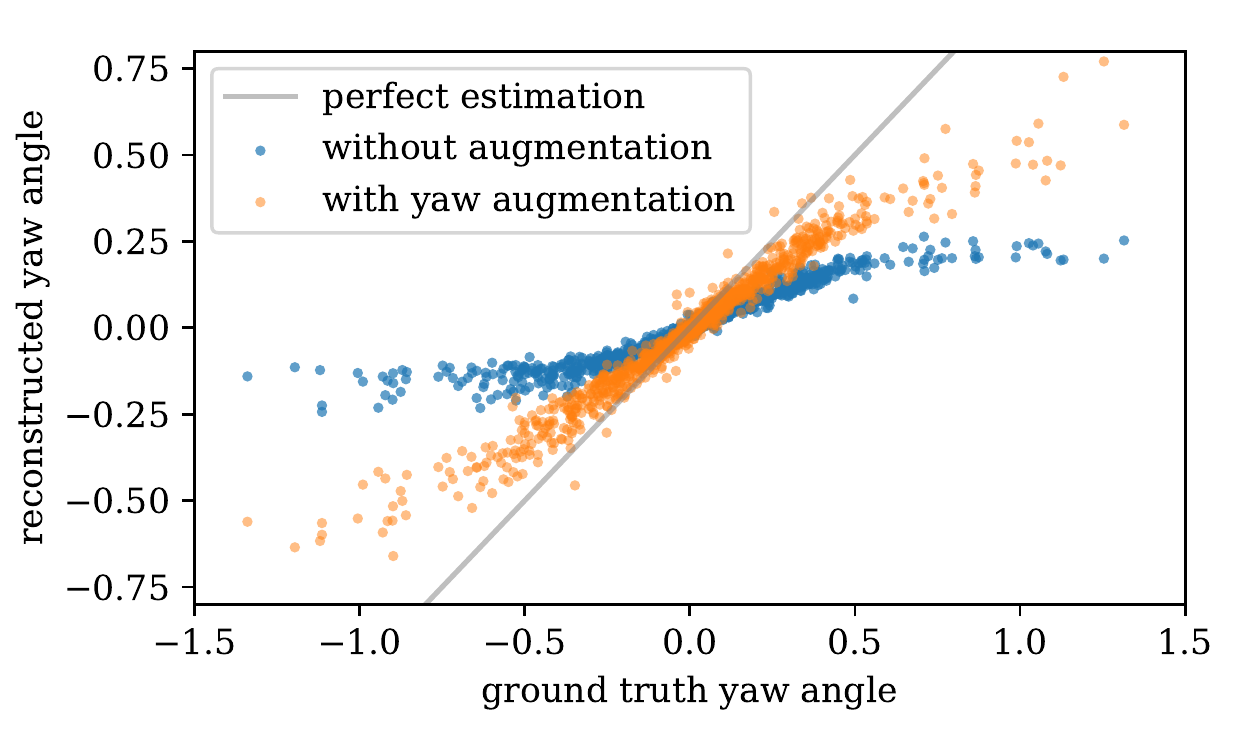}}\label{fig:augmentation_plot}
\subfigure[GAN generations.]{
    \includegraphics[trim={0 3.56cm 0 0.06cm},clip,width=\columnwidth]{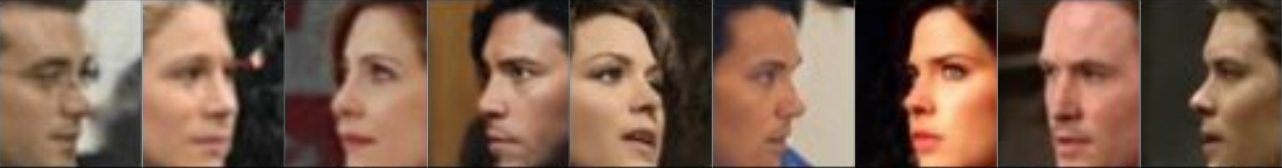}}\\[-0.5ex]
\subfigure[Reconstruction without augmented training.]{
    \includegraphics[trim={0 0.06cm 0 3.56cm},clip,width=\columnwidth]{figs/augment_samples.pdf}}\\[-0.5ex]
\subfigure[Reconstruction with augmented yaw training.]{
    \includegraphics[trim={0 1.8cm 0 1.8cm},clip,width=\columnwidth]{figs/augment_samples.pdf}}
\end{center}
\vspace{-0.7cm}
\caption{The reconstruction result for extreme yaw generations from CASIA StyleGAN. }\label{fig:augmentation_results}
\vspace{-0.3cm}
\end{figure}

\begin{figure*}[ht]
\begin{center}
  \begin{tabular}{@{}c@{}}
    \includegraphics[trim={0 0.74cm 0 0.3cm},clip,width=\textwidth]{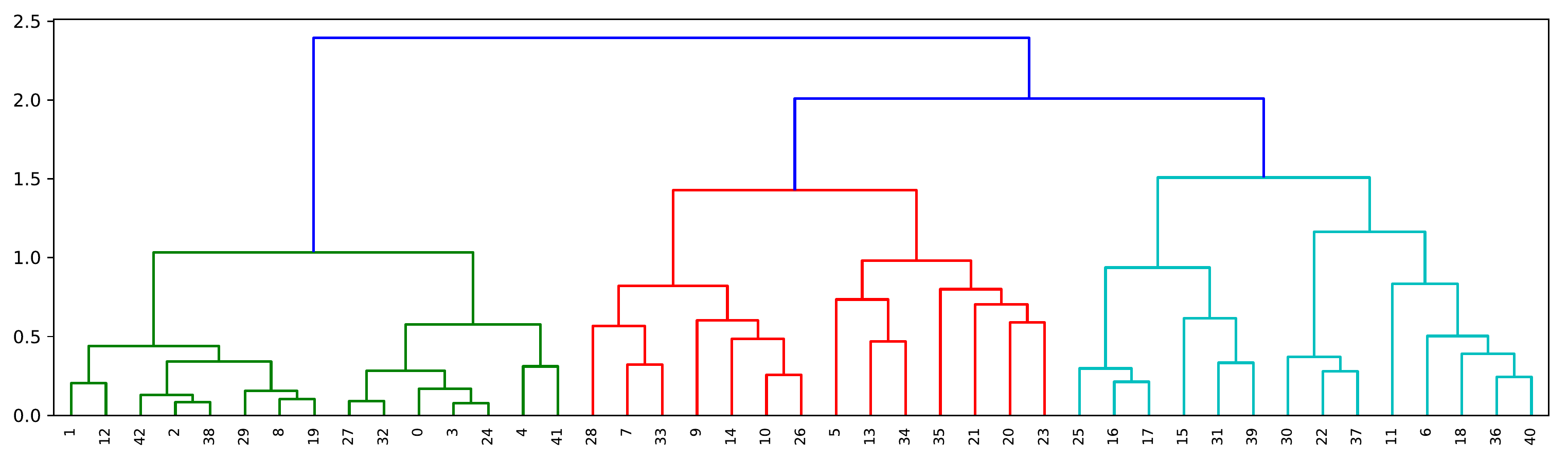}   \\
    \includegraphics[trim={-14cm 0 -5cm 0},width=\textwidth]{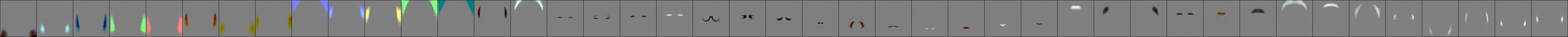} \\
  \end{tabular}
\end{center}
\vspace{-0.7cm}
\caption{The clustering result with Ward variance minimization algorithm \cite{ward1963hierarchical}. The Y-axis stand for the Ward's distance metric calculated from the absolute value of cosine distance. {\color{ForestGreen} \textbf{Green}} cluster: consists of all perturbations in the background. {\color{Red} \textbf{Red}} cluster: contains most facial perturbations inside the face area. {\color{TealBlue} \textbf{Cyan}} cluster: contains mostly facial perturbations around the face region. }\label{fig:canonical_disentangle}
\vspace{-0.3cm}
\end{figure*}

\subsection{Results on univariate semantics manipulation}
In our experiment setting, there are 14 univariate semantics: facial pose (pitch, yaw, roll), face translation (X, Y, Z-axis), ambiance lighting (RGB), diffuse lighting (RGB and direction in X, Y-axis). We apply our linear regression method to get all manipulation vectors for these semantics. Then, the interpolation along each learned manipulation vector is performed with Eqn.~\ref{eqn:manipulate}.

The generation result for StyleGAN trained with FFHQ is shown in Fig.~\ref{fig:univariate_semantics}. 

\textbf{Testing Setup:} To test the behavior of our manipulation vectors, we examine the correlation coefficient between projection length $\mathbf{w}_i^\top \mathbf{v}$ and the ground truth value of the associated semantics. We showcase our result on StyleGAN trained with FFHQ in terms of pitch, yaw angle since the ground truth can be estimated with existing open-source 3D landmarker \cite{bulat2017far}. We ignore roll and all translation semantics because they are almost removed during data pre-processing, and as a result, StyleGAN is not sensitive to these semantics. For lighting semantics, on the other hand, we are currently unable to find any method to accurately estimate their ground truth, thus it is impossible to test their performance quantitatively. Nevertheless, we showcase the visual results of all univariate semantics of both FFHQ and CASIA StyleGAN in the supplementary materials. 

\textbf{Testing Results:} In Table \ref{tab:uni_coeff}, all coefficient values greater than $0.5$ are labeled as bold. It showcases that the pitch and yaw angles are highly correlated with their associated latent representation. Meanwhile, a perturbation in semantically irrelevant latent representations is less likely to affect pitch and yaw. One unexpected phenomenon is the light direction in the X and Y-axis seems to be slightly correlated to yaw and pitch angle, respectively. This is because the viewpoint aligns with the lighting within the same 3D direction in the real world, resulting in a more entangled generation in GANs. The performance drop under extreme lighting conditions is also mentioned in the failure analysis in \cite{wu2020unsupervised}. 

\textbf{Results on self-supervised data augmentation:} We apply yaw angle augmentation on generation from CASIA StyleGAN to improve 3D facial reconstruction, because the generations contain profile or near-profile faces that are hard to reconstruct. The augmentation scale $s$ is a random uniform distributed value with range $[-10,10]$ for extreme extrapolation. We stick to the same optimization settings and compare the profile face matching performance. As shown in Fig~\ref{fig:augmentation_results}, the 3D facial reconstruction algorithm predicts the yaw angle more accurately when trained with yaw augmented generations. We also showcase the reconstruction results of randomly selected profile or near profile examples. 

Due to the success of data augmentation, canonical semantics are ensured to have been better estimated for CASIA StyleGAN generations as well. Thus, for the following canonical semantics, we only report results on the data augmented model for CASIA StyleGAN.

\begin{figure*}[t]
\begin{center}
\includegraphics[trim={0 0 0 0.06cm},clip,width=\textwidth]{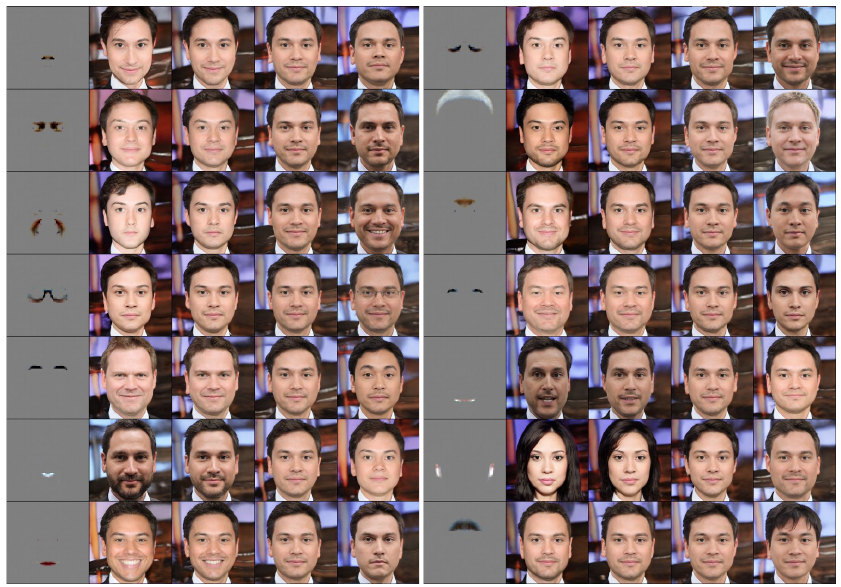}
\end{center}
\vspace{-0.6cm}
\caption{Sampling along the localized representations. A possible interpretation: \textbf{first column}: hooked nose, height of nasal bridge, nasolabial folds, eye glasses, height of eyebrow, mustache/beard, smiling; \textbf{second column}: bags under eyes, darkness of hair, brightness of forehead, depth of orbital rim, opened mouth, long hair, bangs on the forehead. }\label{fig:canonical_manipulation}
\vspace{-0.4cm}
\end{figure*}

\subsection{Results on canonical semantics manipulation}

For our FFHQ experiments, we set $\alpha=1$, $\beta=1$ and $P=200$ in objective~\ref{eqn:sparse_pca} and optimize with a Adam optimizer with learning rate of 0.0001. While for CASIA WebFace experiments, we set $\alpha=5$, $\beta=10$ for best performance. Note that the range for selecting $\alpha$ and $\beta$ is not very strict. We showcase some results by changing the $\alpha$, $\beta$ values in the supplementary materials. The training converges after 500k iterations (25 mins in single RTX 2080 GPU). 

\textbf{Optimization Result:} We discard any $\hat{\mathbf{v}}_p$ if its associated localized facial semantics has $\norm{\mathbf{u}_p}<0.01$, i.e. these $\hat{\mathbf{v}}_p$ does not encode linear semantics in albedo and depth map. Finally, 43 components remain valid for FFHQ StyleGAN. We illustrate their albedo maps as the X-axis in Fig.~\ref{fig:canonical_disentangle}. Among these components, 28 of them are associated with non-background localized facial semantics. In Fig.~\ref{fig:canonical_manipulation}, we demonstrate some example results when performing interpolation/extrapolation along specific localized facial representations $\hat{\mathbf{v}}_p$. The sampling scale $s$ in Eqn.~\ref{eqn:manipulate} ranges from -4 to 4. Note that samples with extreme scale are typically rare for StyleGAN to generate under normal circumstances \cite{shen2020interpreting}. Therefore it is likely for the representation to be less disentangled when extrapolating at a high scale. 

\textbf{Interpretability of Manipulation:} Due to the sparse constraint applied to the localized facial semantics, the resulting $\mathbf{u}_p$ only captures perturbation of limited areas. The sparsity significantly enhances the interpretability of the disentangled facial semantics. For example, in Fig.~\ref{fig:canonical_manipulation}, the semantics is fairly easy to be interpreted.

\textbf{Disentanglement Analysis:} We further check whether these image-level perturbations are semantically encoded in the latent space. The latent representations for semantically independent perturbations are supposed to be more orthogonal with each other; and vice versa. To showcase the result of this test, we decide to use a linkage clustering algorithm on the latent representations $\hat{\mathbf{v}}_p$ and apply Ward's variance minimization \cite{ward1963hierarchical}. The absolute value of the pairwise cosine distance between latent representations is used as the distance metric for the clustering. The result is shown in Fig.~\ref{fig:canonical_disentangle}, where three major clusters are detected. Amazingly, the formation of the clusters is according to the semantics of the facial image, i.e. background areas, facial areas, and areas around the face. Moreover, we find that although the background cluster is tightly distributed, the representations in the other two clusters are fairly independent, as we see the representations in non-background clusters split in a relatively earlier stage in the dendrogram. In general, the localized canonical representations are surprisingly easy to interpret and are complying with human intuitions. 

We also test the semantic correlation of the localized facial semantics and CelebA \cite{liu2015faceattributes} facial attributes. We showcase the similarity by concatenating CelebA SVM weight vectors with $\hat{\mathbf{v}}_p$ and following the same process to generate Fig.~\ref{fig:canonical_disentangle}. The result shows that some semantics are highly correlated with facial attributes, e.g. glasses, smiling, big nose, big lips, etc. Full result is shown in the supplementary materials. 

Besides, CASIA StyleGAN follows a similar trend, where we report in the supplementary materials. 

\section{Conclusion}
We have presented an unsupervised learning framework for disentangling linear-encoded facial semantics from StyleGAN. The system can robustly decompose facial semantics from any single view GAN generations and disentangle facial semantics that is easily interpretable. We also illustrate a potential direction to get the facial manipulation frameworks to work, i.e. performing guided data augmentation to counteract the problem brought by unbalanced data. Extensive analysis suggests that the manipulation of localized facial semantics is easily interpretable and intuitive. 

{\small
\bibliographystyle{ieee_fullname}
\bibliography{egbib}

\begin{thebibliography}{10}\itemsep=-1pt

\bibitem{bao2018towards}
Jianmin Bao, Dong Chen, Fang Wen, Houqiang Li, and Gang Hua.
\newblock Towards open-set identity preserving face synthesis.
\newblock In {\em Proceedings of the IEEE conference on computer vision and
  pattern recognition}, pages 6713--6722, 2018.

\bibitem{blanz1999morphable}
Volker Blanz and Thomas Vetter.
\newblock A morphable model for the synthesis of 3d faces.
\newblock In {\em Proceedings of the 26th annual conference on Computer
  graphics and interactive techniques}, pages 187--194, 1999.

\bibitem{bojanowski2017optimizing}
Piotr Bojanowski, Armand Joulin, David Lopez-Paz, and Arthur Szlam.
\newblock Optimizing the latent space of generative networks.
\newblock {\em arXiv preprint arXiv:1707.05776}, 2017.

\bibitem{bulat2017far}
Adrian Bulat and Georgios Tzimiropoulos.
\newblock How far are we from solving the 2d \& 3d face alignment problem? (and
  a dataset of 230,000 3d facial landmarks).
\newblock In {\em International Conference on Computer Vision}, 2017.

\bibitem{chen2018isolating}
Ricky~TQ Chen, Xuechen Li, Roger~B Grosse, and David~K Duvenaud.
\newblock Isolating sources of disentanglement in variational autoencoders.
\newblock In {\em Advances in Neural Information Processing Systems}, pages
  2610--2620, 2018.

\bibitem{chen2016infogan}
Xi Chen, Yan Duan, Rein Houthooft, John Schulman, Ilya Sutskever, and Pieter
  Abbeel.
\newblock Infogan: Interpretable representation learning by information
  maximizing generative adversarial nets.
\newblock In {\em Advances in neural information processing systems}, pages
  2172--2180, 2016.

\bibitem{choi2018stargan}
Yunjey Choi, Minje Choi, Munyoung Kim, Jung-Woo Ha, Sunghun Kim, and Jaegul
  Choo.
\newblock Stargan: Unified generative adversarial networks for multi-domain
  image-to-image translation.
\newblock In {\em Proceedings of the IEEE conference on computer vision and
  pattern recognition}, pages 8789--8797, 2018.

\bibitem{deng2020disentangled}
Yu Deng, Jiaolong Yang, Dong Chen, Fang Wen, and Xin Tong.
\newblock Disentangled and controllable face image generation via 3d
  imitative-contrastive learning.
\newblock In {\em Proceedings of the IEEE/CVF Conference on Computer Vision and
  Pattern Recognition}, pages 5154--5163, 2020.

\bibitem{desjardins2012disentangling}
Guillaume Desjardins, Aaron Courville, and Yoshua Bengio.
\newblock Disentangling factors of variation via generative entangling.
\newblock {\em arXiv preprint arXiv:1210.5474}, 2012.

\bibitem{donahue2017semantically}
Chris Donahue, Zachary~C Lipton, Akshay Balsubramani, and Julian McAuley.
\newblock Semantically decomposing the latent spaces of generative adversarial
  networks.
\newblock {\em arXiv preprint arXiv:1705.07904}, 2017.

\bibitem{eastwood2018framework}
Cian Eastwood and Christopher~KI Williams.
\newblock A framework for the quantitative evaluation of disentangled
  representations.
\newblock In {\em International Conference on Learning Representations}, 2018.

\bibitem{goodfellow2014generative}
Ian Goodfellow, Jean Pouget-Abadie, Mehdi Mirza, Bing Xu, David Warde-Farley,
  Sherjil Ozair, Aaron Courville, and Yoshua Bengio.
\newblock Generative adversarial nets.
\newblock In {\em Advances in neural information processing systems}, pages
  2672--2680, 2014.

\bibitem{gulrajani2017improved}
Ishaan Gulrajani, Faruk Ahmed, Martin Arjovsky, Vincent Dumoulin, and Aaron~C
  Courville.
\newblock Improved training of wasserstein gans.
\newblock In {\em Advances in neural information processing systems}, pages
  5767--5777, 2017.

\bibitem{heusel2017gans}
Martin Heusel, Hubert Ramsauer, Thomas Unterthiner, Bernhard Nessler, and Sepp
  Hochreiter.
\newblock Gans trained by a two time-scale update rule converge to a local nash
  equilibrium.
\newblock In {\em Advances in neural information processing systems}, pages
  6626--6637, 2017.

\bibitem{higgins2016beta}
Irina Higgins, Loic Matthey, Arka Pal, Christopher Burgess, Xavier Glorot,
  Matthew Botvinick, Shakir Mohamed, and Alexander Lerchner.
\newblock beta-vae: Learning basic visual concepts with a constrained
  variational framework.
\newblock 2016.

\bibitem{jahanian2019steerability}
Ali Jahanian, Lucy Chai, and Phillip Isola.
\newblock On the''steerability" of generative adversarial networks.
\newblock {\em arXiv preprint arXiv:1907.07171}, 2019.

\bibitem{karras2019style}
Tero Karras, Samuli Laine, and Timo Aila.
\newblock A style-based generator architecture for generative adversarial
  networks.
\newblock In {\em Proceedings of the IEEE conference on computer vision and
  pattern recognition}, pages 4401--4410, 2019.

\bibitem{karras2020analyzing}
Tero Karras, Samuli Laine, Miika Aittala, Janne Hellsten, Jaakko Lehtinen, and
  Timo Aila.
\newblock Analyzing and improving the image quality of stylegan.
\newblock In {\em Proceedings of the IEEE/CVF Conference on Computer Vision and
  Pattern Recognition}, pages 8110--8119, 2020.

\bibitem{kim2018disentangling}
Hyunjik Kim and Andriy Mnih.
\newblock Disentangling by factorising.
\newblock {\em arXiv preprint arXiv:1802.05983}, 2018.

\bibitem{kingma2013auto}
Diederik~P Kingma and Max Welling.
\newblock Auto-encoding variational bayes.
\newblock {\em arXiv preprint arXiv:1312.6114}, 2013.

\bibitem{laine2018feature}
Samuli Laine.
\newblock Feature-based metrics for exploring the latent space of generative
  models.
\newblock 2018.

\bibitem{lample2017fader}
Guillaume Lample, Neil Zeghidour, Nicolas Usunier, Antoine Bordes, Ludovic
  Denoyer, and Marc'Aurelio Ranzato.
\newblock Fader networks: Manipulating images by sliding attributes.
\newblock In {\em Advances in neural information processing systems}, pages
  5967--5976, 2017.

\bibitem{lin2019infogan}
Zinan Lin, Kiran~Koshy Thekumparampil, Giulia Fanti, and Sewoong Oh.
\newblock Infogan-cr: Disentangling generative adversarial networks with
  contrastive regularizers.
\newblock {\em arXiv preprint arXiv:1906.06034}, 2019.

\bibitem{liu2019stgan}
Ming Liu, Yukang Ding, Min Xia, Xiao Liu, Errui Ding, Wangmeng Zuo, and Shilei
  Wen.
\newblock Stgan: A unified selective transfer network for arbitrary image
  attribute editing.
\newblock In {\em Proceedings of the IEEE conference on computer vision and
  pattern recognition}, pages 3673--3682, 2019.

\bibitem{liu2015faceattributes}
Ziwei Liu, Ping Luo, Xiaogang Wang, and Xiaoou Tang.
\newblock Deep learning face attributes in the wild.
\newblock In {\em Proceedings of International Conference on Computer Vision
  (ICCV)}, December 2015.

\bibitem{locatello2019challenging}
Francesco Locatello, Stefan Bauer, Mario Lucic, Gunnar Raetsch, Sylvain Gelly,
  Bernhard Sch{\"o}lkopf, and Olivier Bachem.
\newblock Challenging common assumptions in the unsupervised learning of
  disentangled representations.
\newblock In {\em international conference on machine learning}, pages
  4114--4124. PMLR, 2019.

\bibitem{martin2017wasserstein}
SC Martin~Arjovsky and Leon Bottou.
\newblock Wasserstein generative adversarial networks.
\newblock In {\em Proceedings of the 34 th International Conference on Machine
  Learning, Sydney, Australia}, 2017.

\bibitem{miyato2018spectral}
Takeru Miyato, Toshiki Kataoka, Masanori Koyama, and Yuichi Yoshida.
\newblock Spectral normalization for generative adversarial networks.
\newblock {\em arXiv preprint arXiv:1802.05957}, 2018.

\bibitem{odena2017conditional}
Augustus Odena, Christopher Olah, and Jonathon Shlens.
\newblock Conditional image synthesis with auxiliary classifier gans.
\newblock In {\em International conference on machine learning}, pages
  2642--2651, 2017.

\bibitem{rezende2014stochastic}
Danilo~Jimenez Rezende, Shakir Mohamed, and Daan Wierstra.
\newblock Stochastic backpropagation and approximate inference in deep
  generative models.
\newblock {\em arXiv preprint arXiv:1401.4082}, 2014.

\bibitem{shao2018riemannian}
Hang Shao, Abhishek Kumar, and P Thomas~Fletcher.
\newblock The riemannian geometry of deep generative models.
\newblock In {\em Proceedings of the IEEE Conference on Computer Vision and
  Pattern Recognition Workshops}, pages 315--323, 2018.

\bibitem{shen2020interpreting}
Yujun Shen, Jinjin Gu, Xiaoou Tang, and Bolei Zhou.
\newblock Interpreting the latent space of gans for semantic face editing.
\newblock In {\em Proceedings of the IEEE/CVF Conference on Computer Vision and
  Pattern Recognition}, pages 9243--9252, 2020.

\bibitem{shen2018faceid}
Yujun Shen, Ping Luo, Junjie Yan, Xiaogang Wang, and Xiaoou Tang.
\newblock Faceid-gan: Learning a symmetry three-player gan for
  identity-preserving face synthesis.
\newblock In {\em Proceedings of the IEEE conference on computer vision and
  pattern recognition}, pages 821--830, 2018.

\bibitem{shen2020interfacegan}
Yujun Shen, Ceyuan Yang, Xiaoou Tang, and Bolei Zhou.
\newblock Interfacegan: Interpreting the disentangled face representation
  learned by gans.
\newblock {\em arXiv preprint arXiv:2005.09635}, 2020.

\bibitem{shen2018facefeat}
Yujun Shen, Bolei Zhou, Ping Luo, and Xiaoou Tang.
\newblock Facefeat-gan: a two-stage approach for identity-preserving face
  synthesis.
\newblock {\em arXiv preprint arXiv:1812.01288}, 2018.

\bibitem{tran2017disentangled}
Luan Tran, Xi Yin, and Xiaoming Liu.
\newblock Disentangled representation learning gan for pose-invariant face
  recognition.
\newblock In {\em Proceedings of the IEEE conference on computer vision and
  pattern recognition}, pages 1415--1424, 2017.

\bibitem{vahdat2020nvae}
Arash Vahdat and Jan Kautz.
\newblock Nvae: A deep hierarchical variational autoencoder.
\newblock {\em Advances in Neural Information Processing Systems}, 33, 2020.

\bibitem{van2017neural}
Aaron Van Den~Oord, Oriol Vinyals, et~al.
\newblock Neural discrete representation learning.
\newblock In {\em Advances in Neural Information Processing Systems}, pages
  6306--6315, 2017.

\bibitem{ward1963hierarchical}
Joe~H Ward~Jr.
\newblock Hierarchical grouping to optimize an objective function.
\newblock {\em Journal of the American statistical association},
  58(301):236--244, 1963.

\bibitem{wu2020unsupervised}
Shangzhe Wu, Christian Rupprecht, and Andrea Vedaldi.
\newblock Unsupervised learning of probably symmetric deformable 3d objects
  from images in the wild.
\newblock In {\em Proceedings of the IEEE/CVF Conference on Computer Vision and
  Pattern Recognition}, pages 1--10, 2020.

\bibitem{yang2019semantic}
Ceyuan Yang, Yujun Shen, and Bolei Zhou.
\newblock Semantic hierarchy emerges in deep generative representations for
  scene synthesis.
\newblock {\em arXiv preprint arXiv:1911.09267}, 2019.

\bibitem{yi2014learning}
Dong Yi, Zhen Lei, Shengcai Liao, and Stan~Z Li.
\newblock Learning face representation from scratch.
\newblock {\em arXiv preprint arXiv:1411.7923}, 2014.

\bibitem{yin2017towards}
Xi Yin, Xiang Yu, Kihyuk Sohn, Xiaoming Liu, and Manmohan Chandraker.
\newblock Towards large-pose face frontalization in the wild.
\newblock In {\em Proceedings of the IEEE international conference on computer
  vision}, pages 3990--3999, 2017.

\bibitem{zhang2019self}
Han Zhang, Ian Goodfellow, Dimitris Metaxas, and Augustus Odena.
\newblock Self-attention generative adversarial networks.
\newblock In {\em International Conference on Machine Learning}, pages
  7354--7363. PMLR, 2019.

\bibitem{zhu2017unpaired}
Jun-Yan Zhu, Taesung Park, Phillip Isola, and Alexei~A Efros.
\newblock Unpaired image-to-image translation using cycle-consistent
  adversarial networks.
\newblock In {\em Proceedings of the IEEE international conference on computer
  vision}, pages 2223--2232, 2017.

\end{thebibliography}
}

\end{document}

\maketitle
\thispagestyle{empty}

\vspace{-3.6cm}
\section{Additional results on univariate semantics}
\begin{figure}[H]
\centering
\includegraphics[trim={0 0 0 0},clip,width=0.9\textwidth]{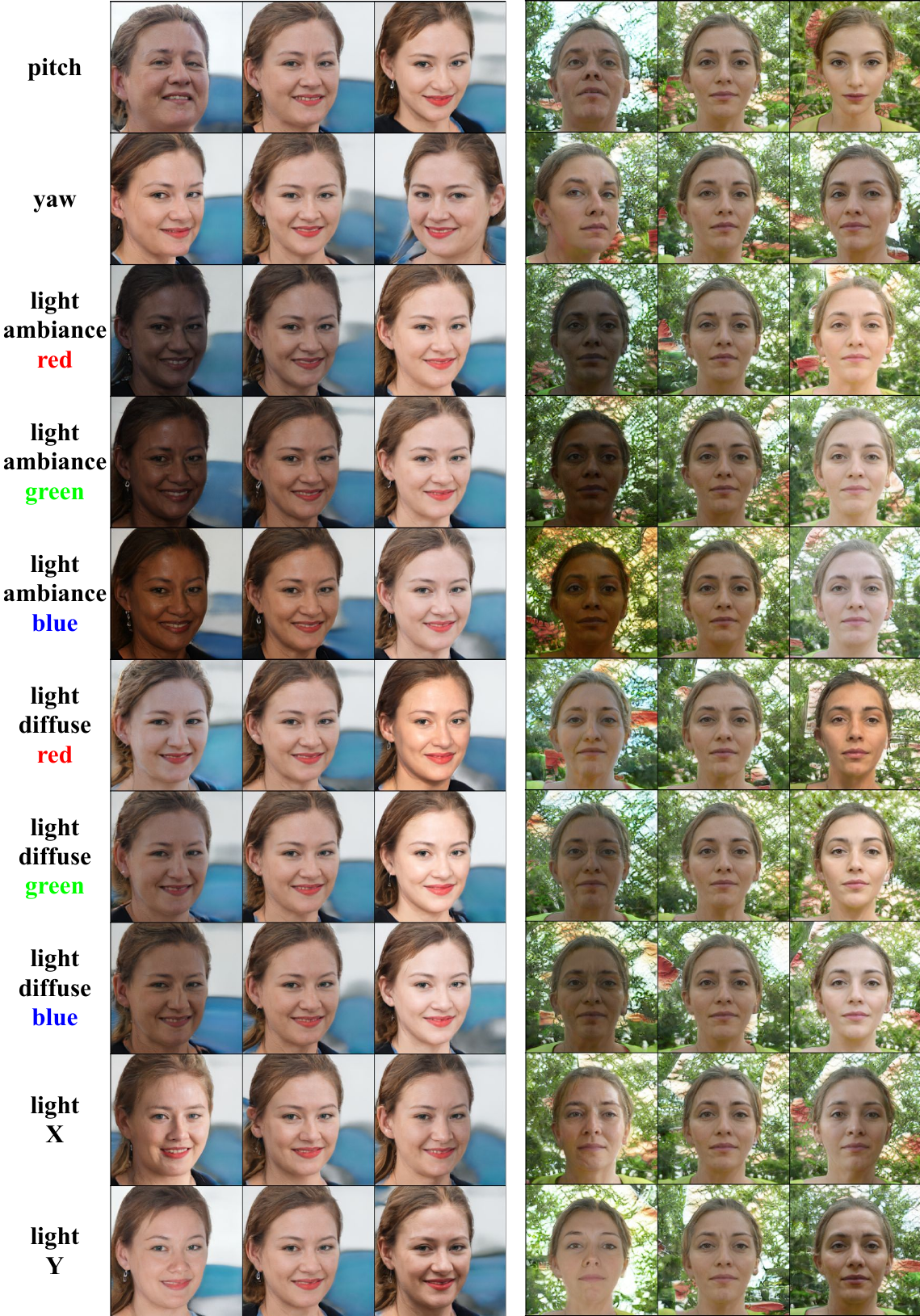}
% \vspace{-0.6cm}
\caption{FFHQ samples.}
% \vspace{-0.4cm}
\end{figure}

\newpage

\begin{figure}[H]
\begin{center}
\includegraphics[trim={0 0 0 0.},clip,width=0.9\textwidth]{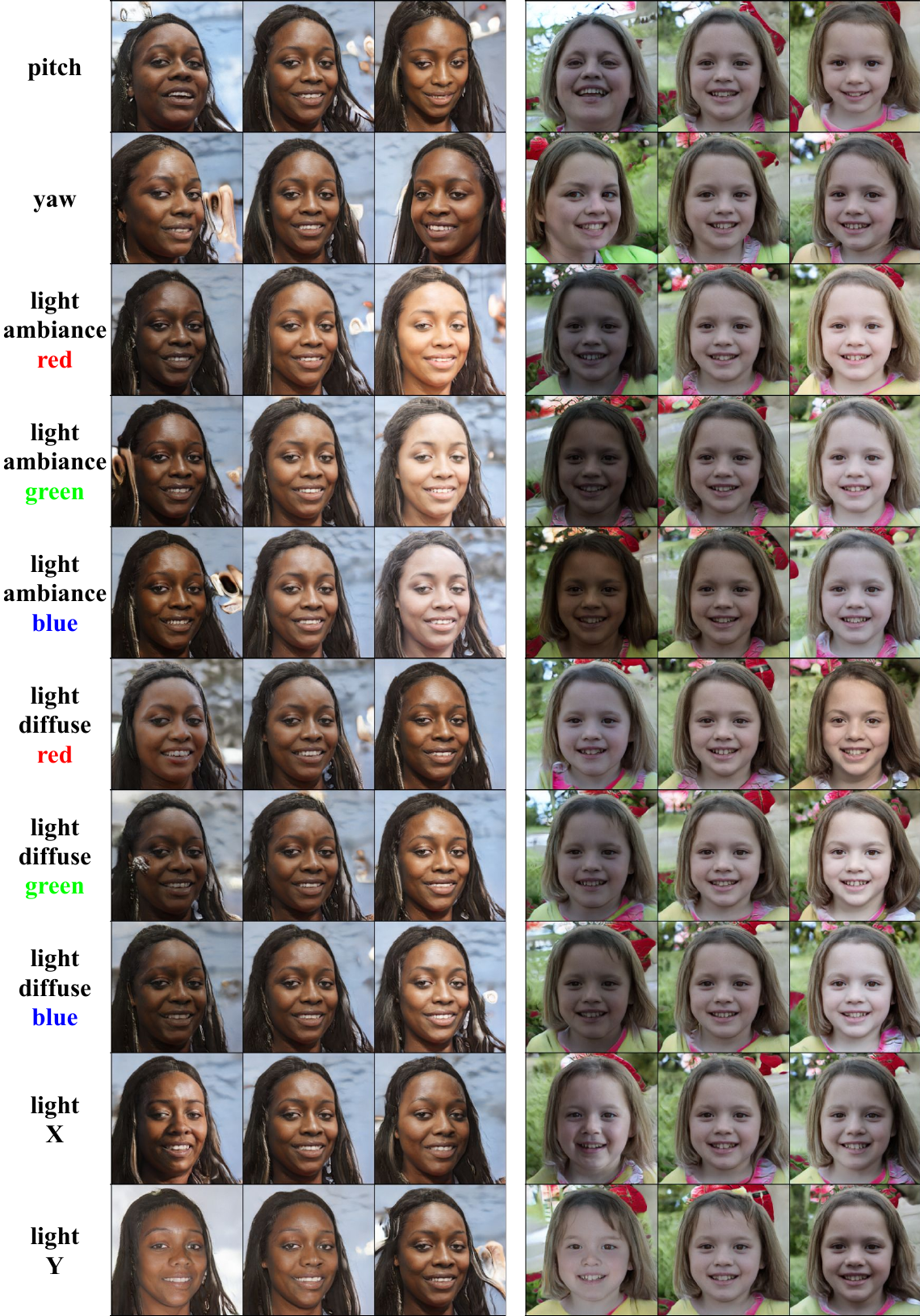}
\end{center}
\vspace{-0.6cm}
\caption{FFHQ samples.}
\vspace{-0.4cm}
\end{figure}

\newpage

\begin{figure}[H]
\begin{center}
\includegraphics[trim={0 0 0 0.},clip,width=0.9\textwidth]{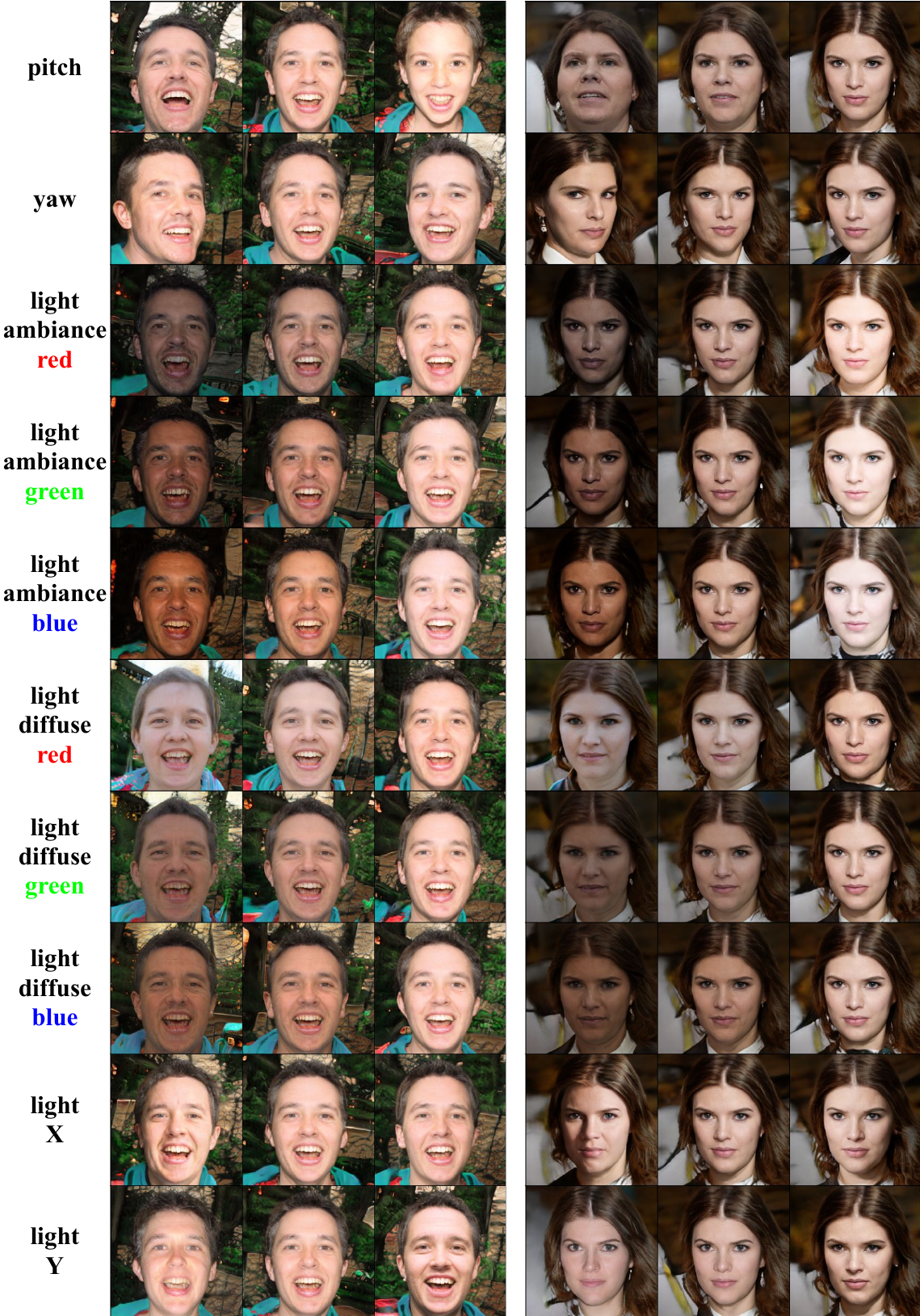}
\end{center}
\vspace{-0.6cm}
\caption{FFHQ samples.}
\vspace{-0.4cm}
\end{figure}

\newpage

\begin{figure}[h]
\begin{center}
\includegraphics[trim={0 0 0 0.},clip,width=0.8\textwidth]{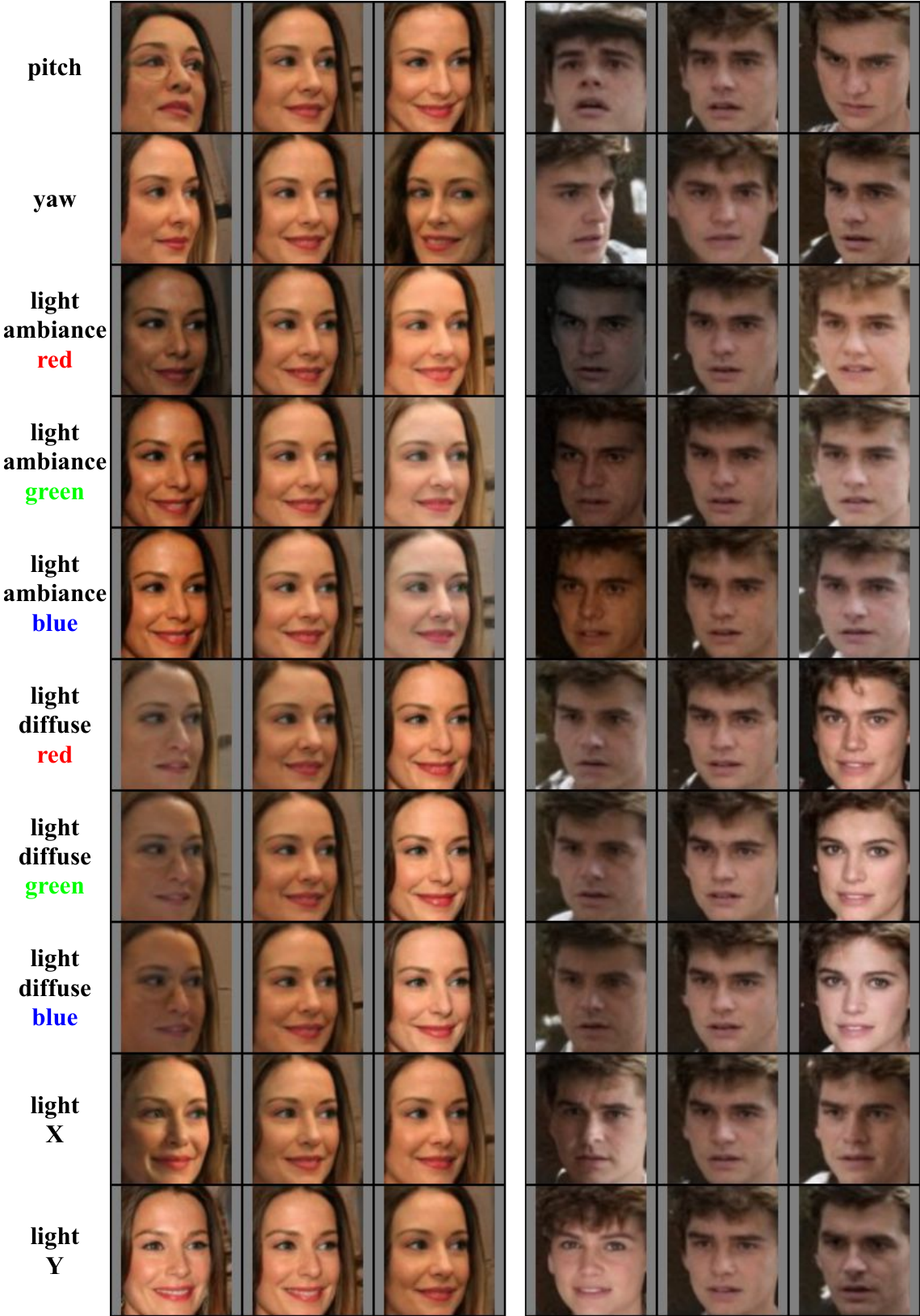}
\end{center}
\vspace{-0.6cm}
\caption{CASIA samples.}
\vspace{-0.4cm}
\end{figure}

\newpage

\begin{figure}[h]
\begin{center}
\includegraphics[trim={0 0 0 0.},clip,width=0.8\textwidth]{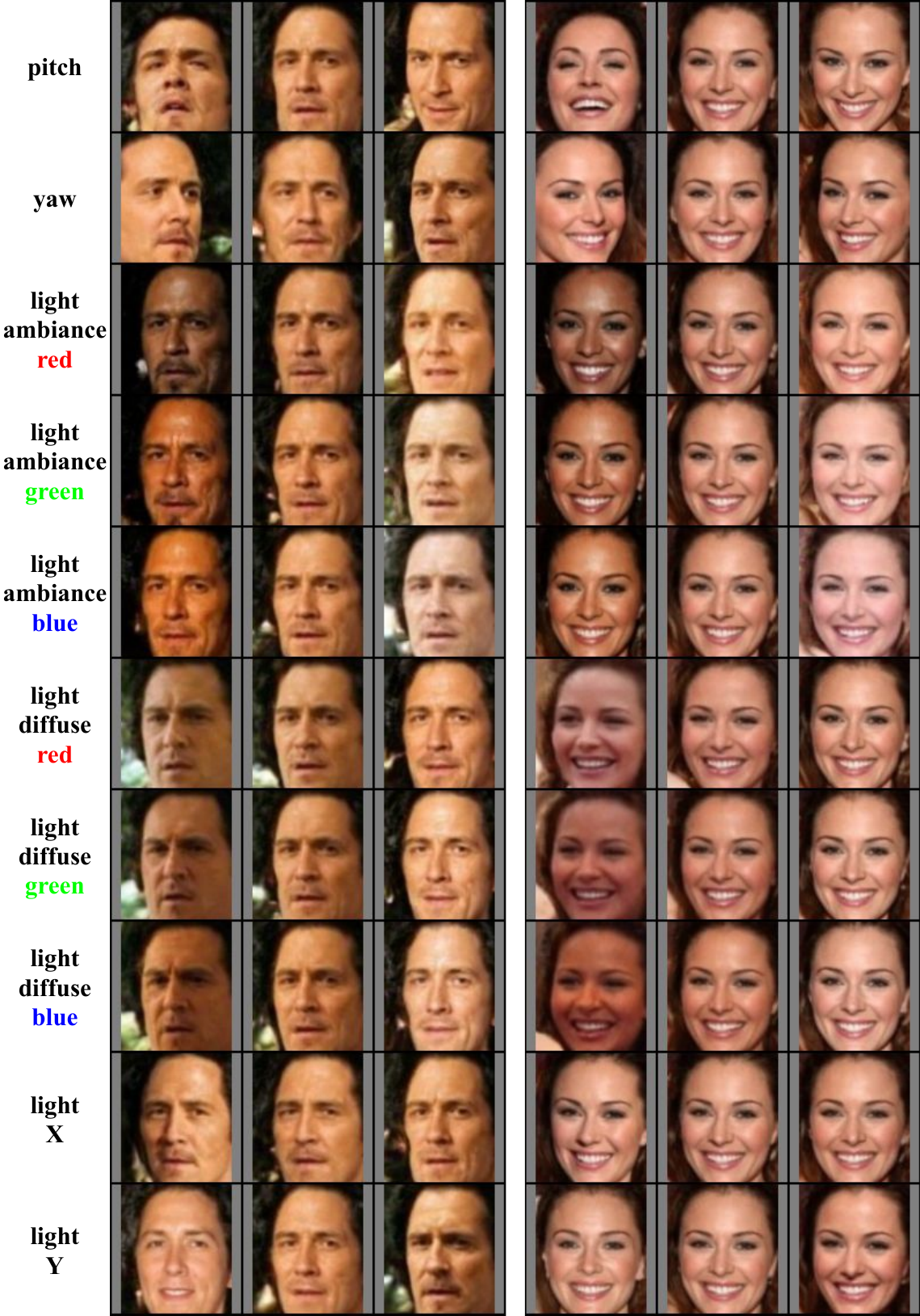}
\end{center}
\vspace{-0.6cm}
\caption{CASIA samples.}
\vspace{-0.4cm}
\end{figure}

\newpage

\begin{figure}[h]
\begin{center}
\includegraphics[trim={0 0 0 0.},clip,width=0.8\textwidth]{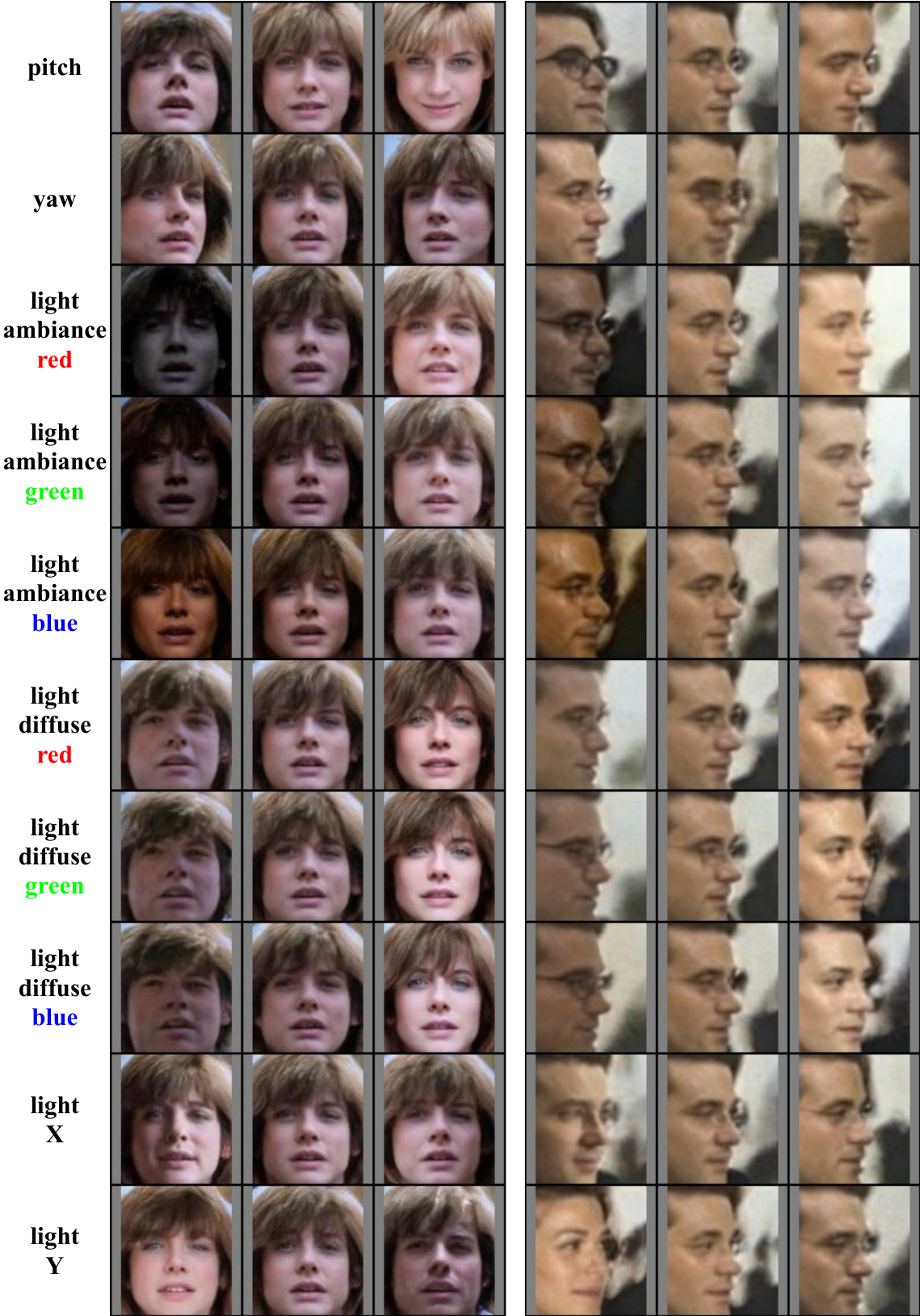}
\end{center}
\vspace{-0.6cm}
\caption{CASIA samples.}
\vspace{-0.4cm}
\end{figure}

\newpage

\section{Additional results on canonical semantics manipulation}

\begin{figure}[H]
\begin{center}
\includegraphics[trim={0 0 0 0.},clip,width=0.8\textwidth]{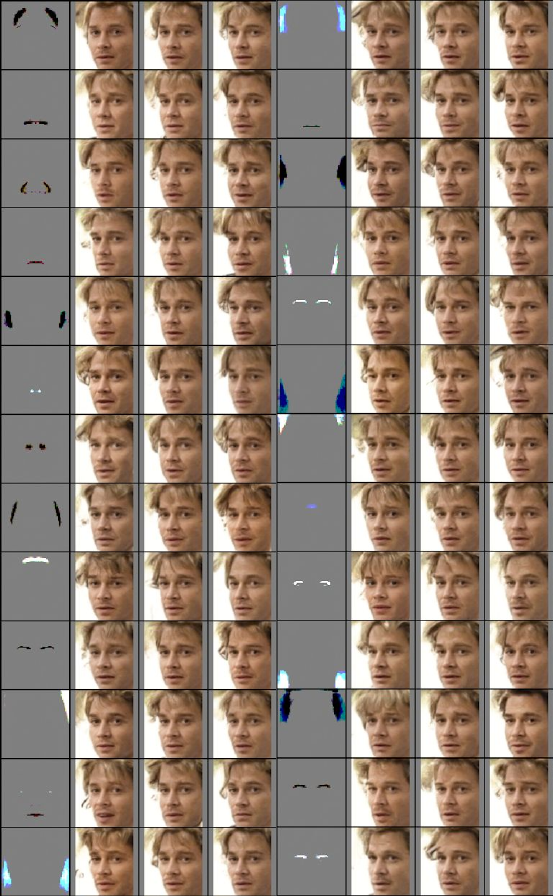}
\end{center}
\vspace{-0.6cm}
\caption{CASIA samples.}
\vspace{-0.4cm}
\end{figure}

\newpage

\begin{figure}[H]
\begin{center}
\includegraphics[trim={0 0 0 0.},clip,width=0.8\textwidth]{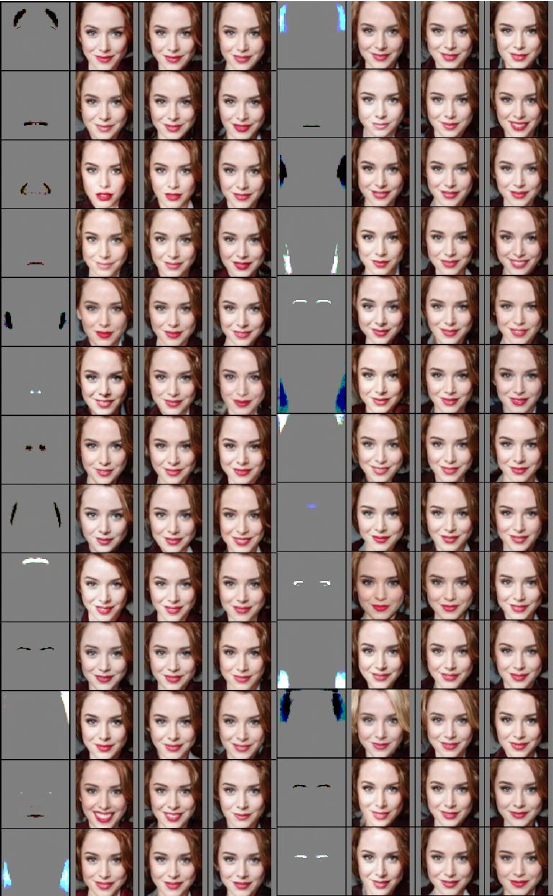}
\end{center}
\vspace{-0.6cm}
\caption{CASIA samples.}
\vspace{-0.4cm}
\end{figure}

\newpage

\begin{figure}[H]
\begin{center}
\includegraphics[trim={0 0 0 0.},clip,width=0.8\textwidth]{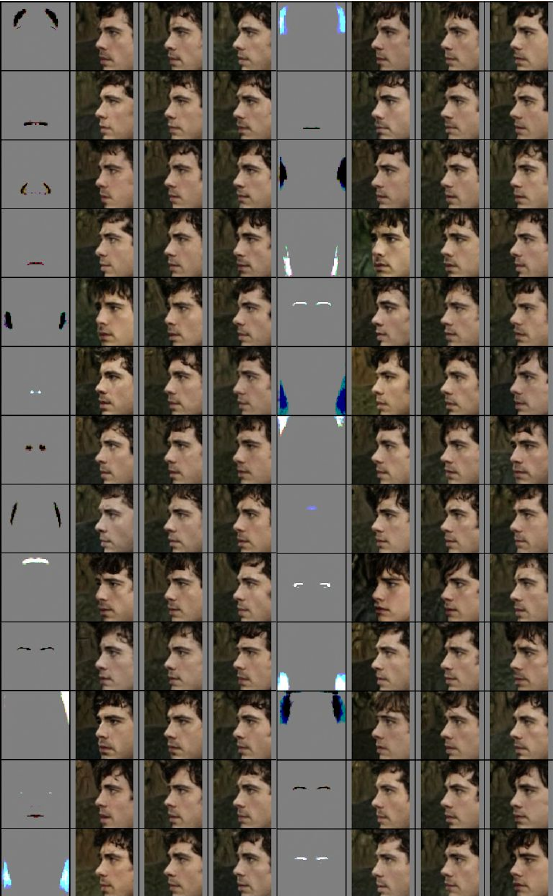}
\end{center}
\vspace{-0.6cm}
\caption{CASIA samples.}
\vspace{-0.4cm}
\end{figure}

\newpage

\begin{figure}[H]
\begin{center}
\includegraphics[trim={0 0 0 0.},clip,width=0.8\textwidth]{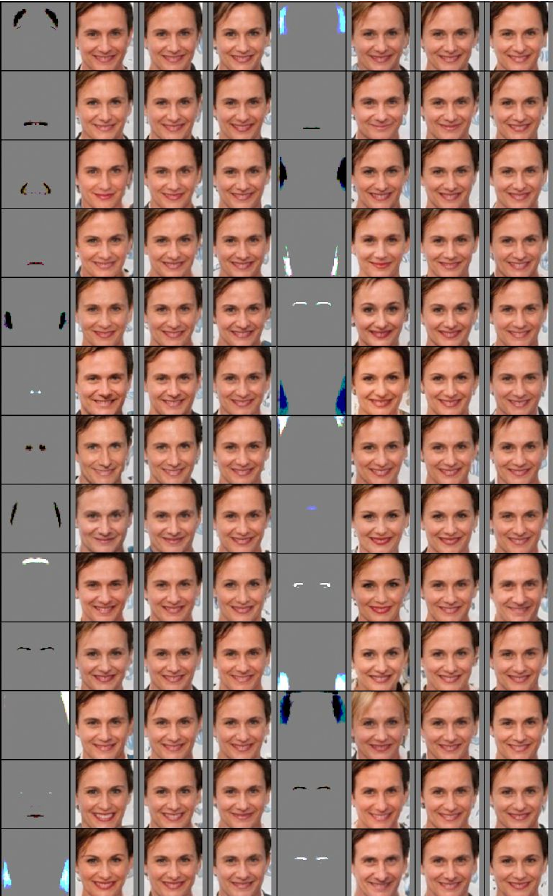}
\end{center}
\vspace{-0.6cm}
\caption{CASIA samples.}
\vspace{-0.4cm}
\end{figure}

\newpage

\begin{figure}[H]
\begin{center}
\includegraphics[trim={0 0 0 0.},clip,width=0.8\textwidth]{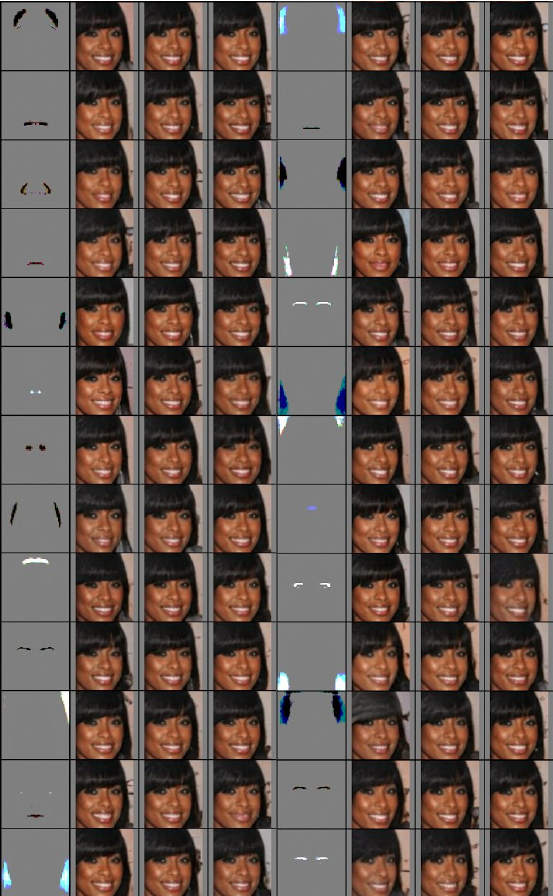}
\end{center}
\vspace{-0.6cm}
\caption{CASIA samples.}
\vspace{-0.4cm}
\end{figure}

\newpage

\begin{figure}[H]
\begin{center}
\includegraphics[trim={0 0 0 0.},clip,width=\textwidth]{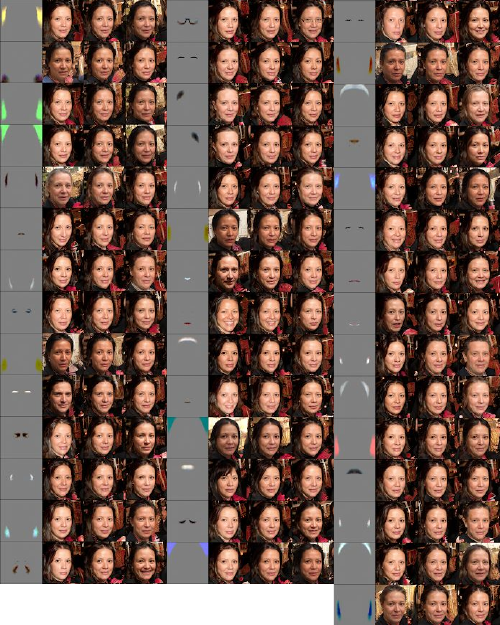}
\end{center}
\vspace{-0.6cm}
\caption{FFHQ samples.}
\vspace{-0.4cm}
\end{figure}

\newpage

\begin{figure}[H]
\begin{center}
\includegraphics[trim={0 0 0 0.},clip,width=\textwidth]{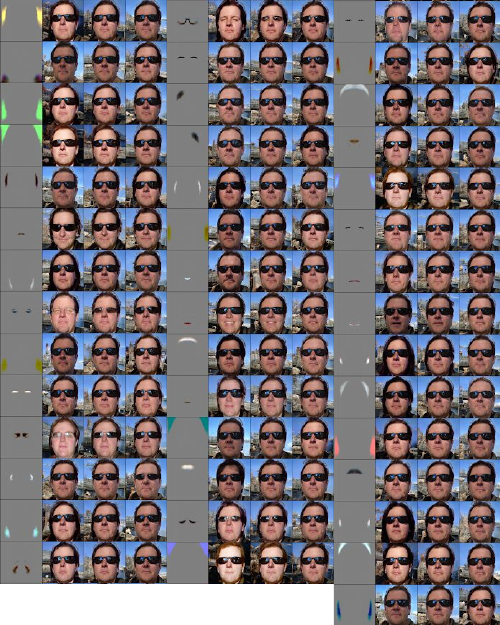}
\end{center}
\vspace{-0.6cm}
\caption{FFHQ samples.}
\vspace{-0.4cm}
\end{figure}

\newpage

\begin{figure}[H]
\begin{center}
\includegraphics[trim={0 0 0 0.},clip,width=\textwidth]{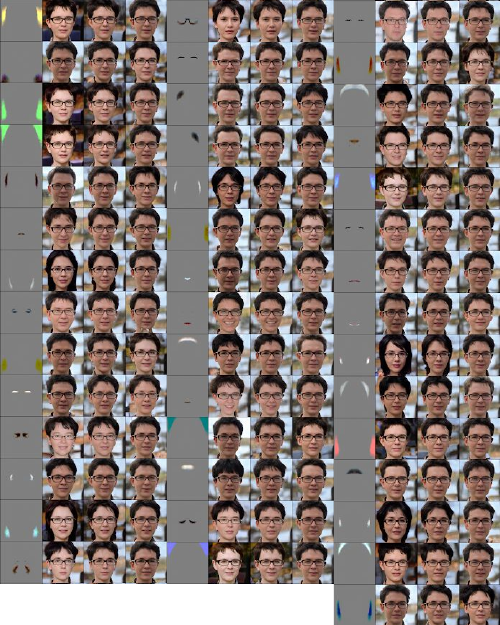}
\end{center}
\vspace{-0.6cm}
\caption{FFHQ samples.}
\vspace{-0.4cm}
\end{figure}

\newpage

\begin{figure}[H]
\begin{center}
\includegraphics[trim={0 0 0 0.},clip,width=\textwidth]{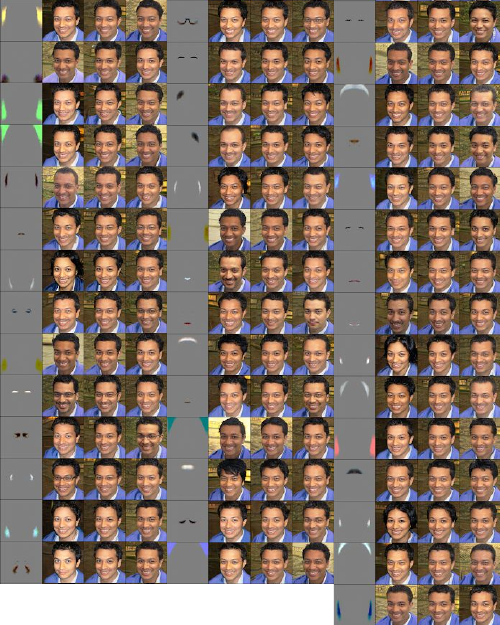}
\end{center}
\vspace{-0.6cm}
\caption{FFHQ samples.}
\vspace{-0.4cm}
\end{figure}

\newpage

\begin{figure}[H]
\begin{center}
\includegraphics[trim={0 0 0 0.},clip,width=\textwidth]{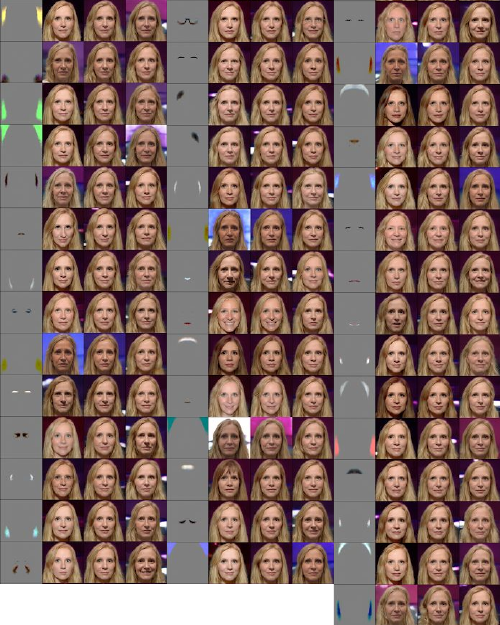}
\end{center}
\vspace{-0.6cm}
\caption{FFHQ samples.}
\vspace{-0.4cm}
\end{figure}

\newpage

\subsection{Changing $\alpha$ value}
By increasing $\alpha$, the number of canonical components decreases and become more sparse. However, they are not necessarily more independent. 

\begin{figure}[H]
\begin{center}
\subfigure{
        \includegraphics[ width=\columnwidth]{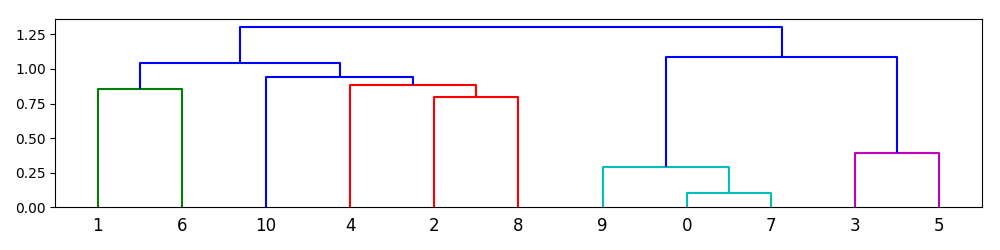}}
\subfigure[{\color{ForestGreen} \textbf{Green}}]{
        \includegraphics[ width=0.2\columnwidth]{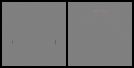}}
\subfigure[{\color{blue} \textbf{Blue}}]{
        \includegraphics[ width=0.1\columnwidth]{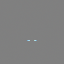}}
\subfigure[{\color{Red} \textbf{Red}}]{
        \includegraphics[ width=0.3\columnwidth]{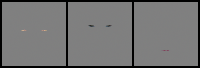}}
\subfigure[{\color{TealBlue} \textbf{Cyan}}]{
        \includegraphics[ width=0.3\columnwidth]{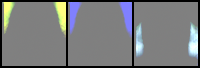}}
\subfigure[{\color{Purple} \textbf{Purple}}]{
        \includegraphics[ width=0.2\columnwidth]{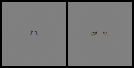}}
\end{center}
\vspace{-0.6cm}
\caption{Albedo components of FFHQ StyleGAN with $\alpha=3$, $\beta=1$.}
\vspace{-0.4cm}
\end{figure}

\newpage

\begin{figure}[H]
\begin{center}
\subfigure{
        \includegraphics[ width=\columnwidth]{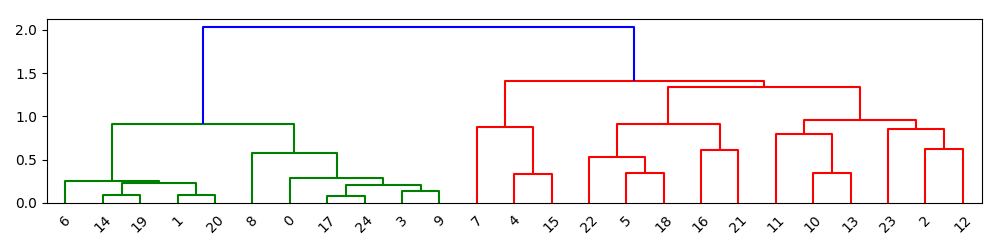}}
\subfigure[{\color{ForestGreen} \textbf{Green}}]{
        \includegraphics[ width=0.9\columnwidth]{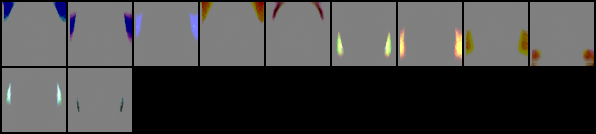}}
\subfigure[{\color{Red} \textbf{Red}}]{
        \includegraphics[ width=0.9\columnwidth]{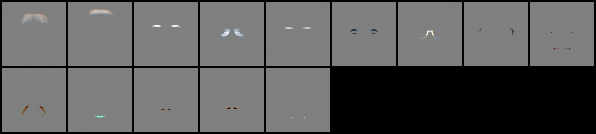}}
\end{center}
\vspace{-0.6cm}
\caption{Albedo components of FFHQ StyleGAN with $\alpha=2$, $\beta=1$.}
\vspace{-0.4cm}
\end{figure}

\newpage

\begin{figure}[H]
\begin{center}
\subfigure{
        \includegraphics[ width=\columnwidth]{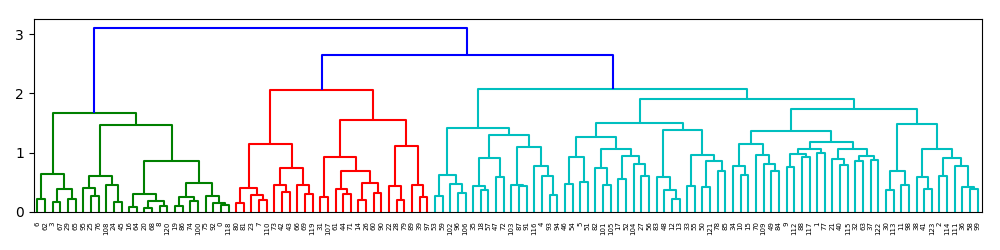}}
\subfigure[{\color{ForestGreen} \textbf{Green}}]{
        \includegraphics[ width=0.7\columnwidth]{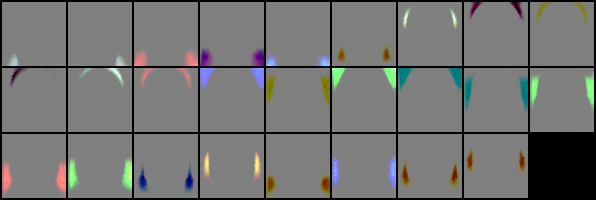}}
\subfigure[{\color{Red} \textbf{Red}}]{
        \includegraphics[ width=0.7\columnwidth]{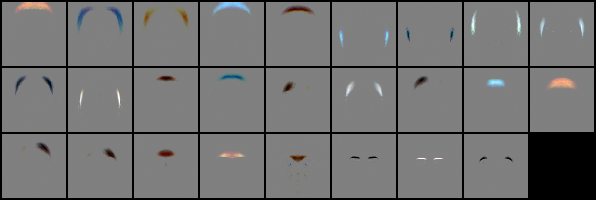}}
\subfigure[{\color{TealBlue} \textbf{Cyan}}]{
        \includegraphics[ width=0.7\columnwidth]{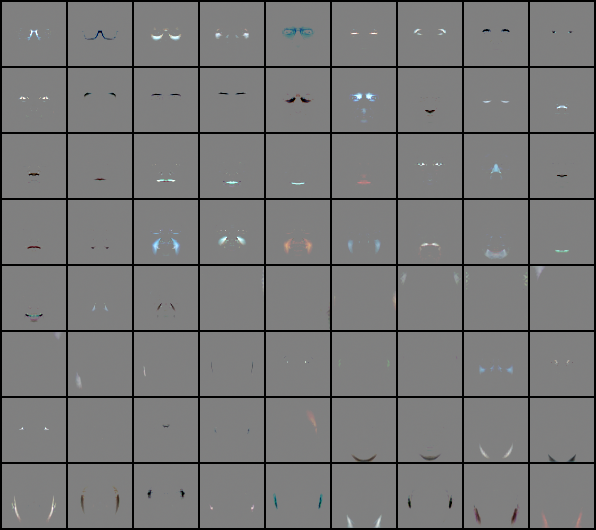}}
\end{center}
\vspace{-0.6cm}
\caption{Albedo components of FFHQ StyleGAN with $\alpha=0.3$, $\beta=1$.}
\vspace{-0.4cm}
\end{figure}

\newpage

\subsection{Changing $\beta$ value}
By increasing $\beta$, we see the components distributed more independently and tend to control more pixels in the image domain. 

\begin{figure}[H]
\begin{center}
\subfigure{
        \includegraphics[ width=\columnwidth]{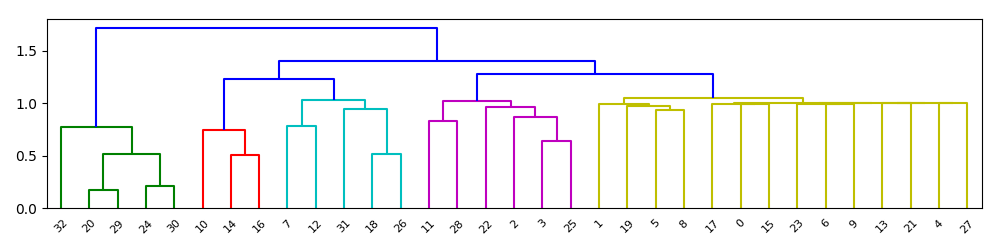}}
\subfigure[{\color{ForestGreen} \textbf{Green}}]{
        \includegraphics[ width=0.5\columnwidth]{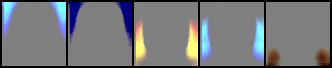}}
\subfigure[{\color{Red} \textbf{Red}}]{
        \includegraphics[ width=0.3\columnwidth]{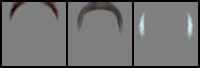}}
\subfigure[{\color{TealBlue} \textbf{Cyan}}]{
        \includegraphics[ width=0.5\columnwidth]{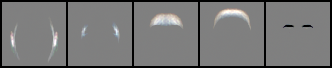}}
\subfigure[{\color{Purple} \textbf{Purple}}]{
        \includegraphics[ width=0.6\columnwidth]{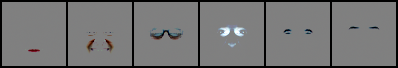}}
\subfigure[{\color{SpringGreen} \textbf{Yellow}}]{
        \includegraphics[ width=0.9\columnwidth]{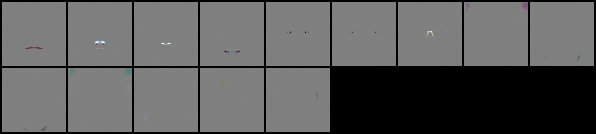}}
\end{center}
\vspace{-0.6cm}
\caption{Albedo components of FFHQ StyleGAN with $\alpha=1$, $\beta=100$.}
\vspace{-0.4cm}
\end{figure}

\newpage

\begin{figure}[H]
\begin{center}
\subfigure{
        \includegraphics[ width=\columnwidth]{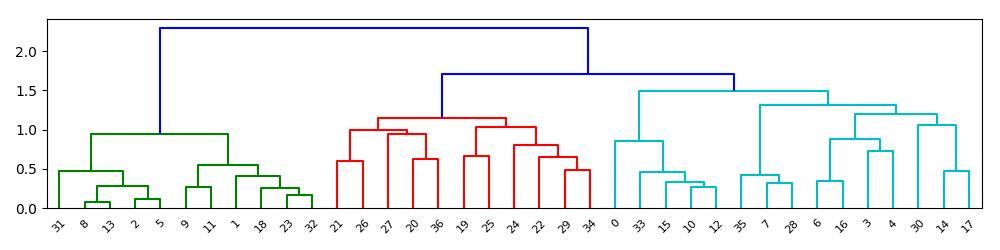}}
\subfigure[{\color{ForestGreen} \textbf{Green}}]{
        \includegraphics[ width=0.9\columnwidth]{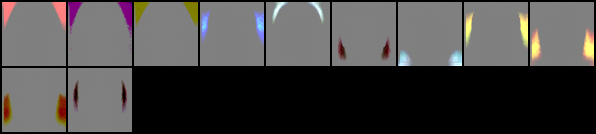}}
\subfigure[{\color{Red} \textbf{Red}}]{
        \includegraphics[ width=0.9\columnwidth]{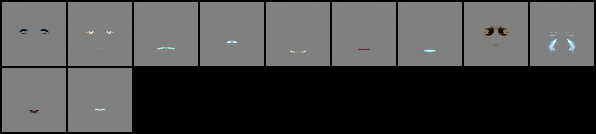}}
\subfigure[{\color{TealBlue} \textbf{Cyan}}]{
        \includegraphics[ width=0.9\columnwidth]{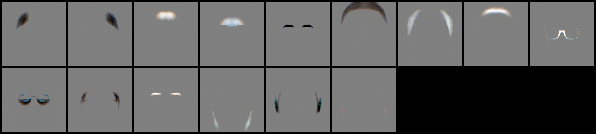}}
\end{center}
\vspace{-0.6cm}
\caption{Albedo components of FFHQ StyleGAN with $\alpha=1$, $\beta=10$.}
\vspace{-0.4cm}
\end{figure}

\newpage

\begin{figure}[H]
\begin{center}
\subfigure{
        \includegraphics[ width=\columnwidth]{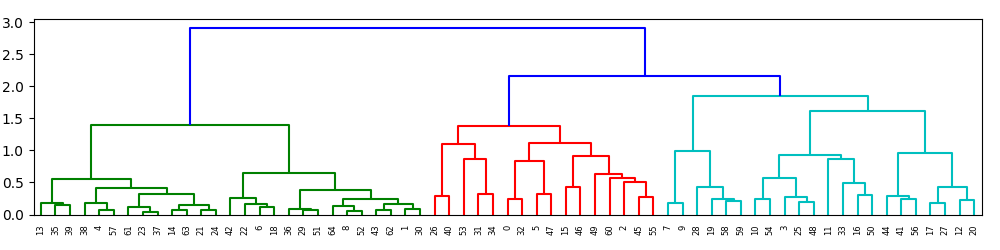}}
\subfigure[{\color{ForestGreen} \textbf{Green}}]{
        \includegraphics[ width=0.9\columnwidth]{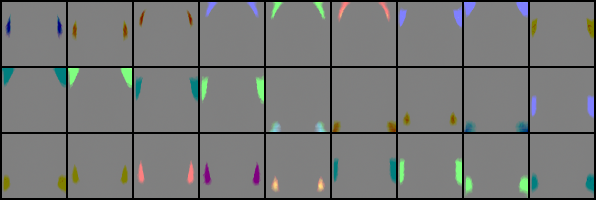}}
\subfigure[{\color{Red} \textbf{Red}}]{
        \includegraphics[ width=0.9\columnwidth]{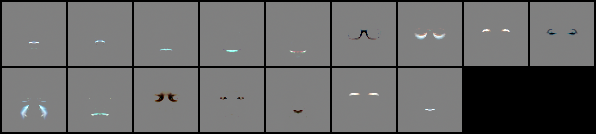}}
\subfigure[{\color{TealBlue} \textbf{Cyan}}]{
        \includegraphics[ width=0.9\columnwidth]{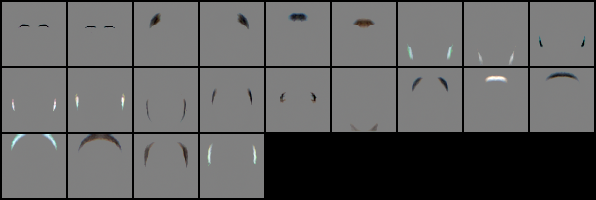}}
\end{center}
\vspace{-0.6cm}
\caption{Albedo components of FFHQ StyleGAN with $\alpha=1$, $\beta=0.1$.}
\vspace{-0.4cm}
\end{figure}

\newpage

\begin{figure}[H]
\begin{center}
\subfigure{
        \includegraphics[ width=\columnwidth]{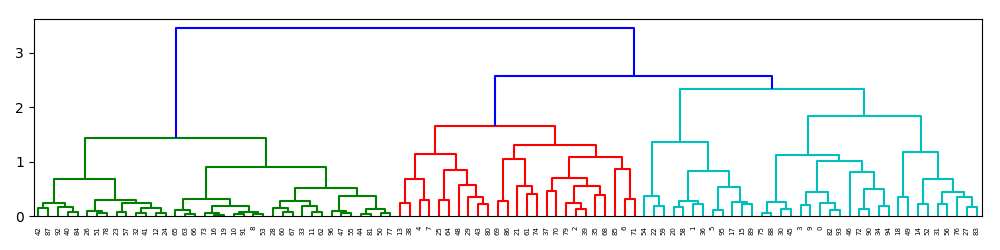}}
\subfigure[{\color{ForestGreen} \textbf{Green}}]{
        \includegraphics[ width=0.8\columnwidth]{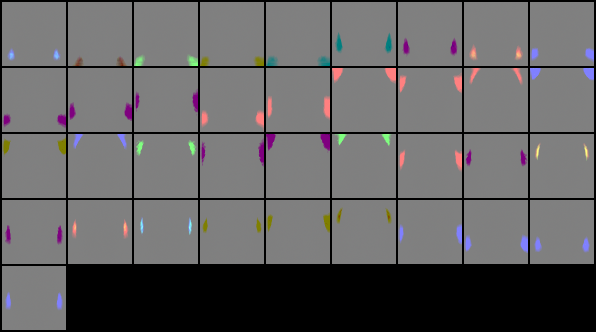}}
\subfigure[{\color{Red} \textbf{Red}}]{
        \includegraphics[ width=0.8\columnwidth]{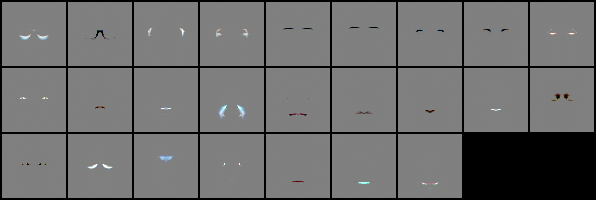}}
\subfigure[{\color{TealBlue} \textbf{Cyan}}]{
        \includegraphics[ width=0.8\columnwidth]{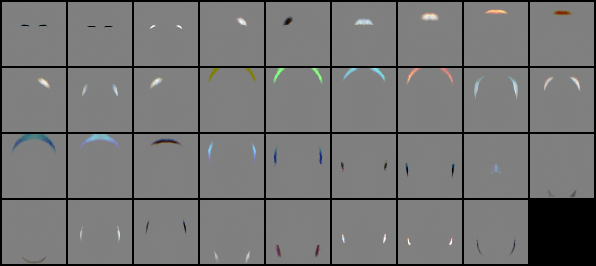}}
\end{center}
\vspace{-0.6cm}
\caption{Albedo components of FFHQ StyleGAN with $\alpha=1$, $\beta=0.01$.}
\vspace{-0.4cm}
\end{figure}

\newpage

\subsection{Similarity with CelebA facial attributes}
\begin{figure}[H]
\begin{center}
\subfigure{
        \includegraphics[ width=\columnwidth]{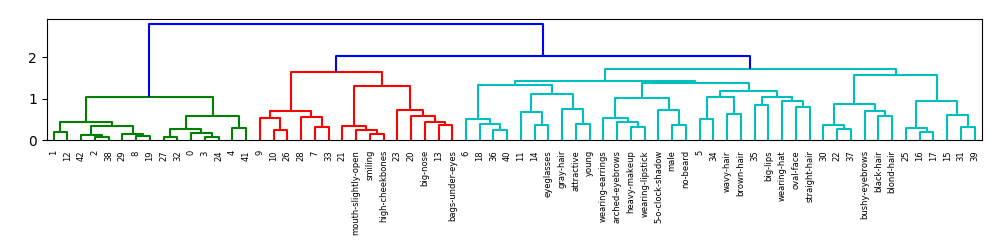}}
\subfigure[{\color{ForestGreen} \textbf{Green}: 1, 12, 2, 38, 42, 8, 19, 29, 27, 32, 3, 24, 0, 4, 41}]{
        \includegraphics[ width=0.8\columnwidth]{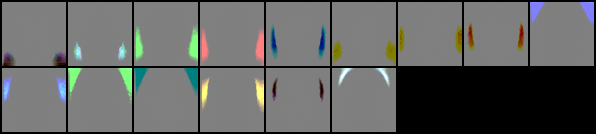}}
\subfigure[{\color{Red} \textbf{Red}: 10, 26, 9, 7, 33, 28, 21, 13, 20, 23}]{
        \includegraphics[ width=0.8\columnwidth]{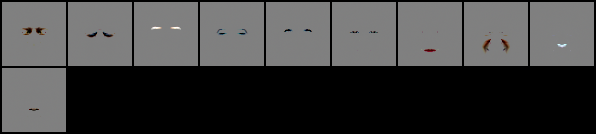}}
\subfigure[{\color{TealBlue} \textbf{Cyan}: 36, 40, 18, 6, 14, 11, 5, 34, 35, 22, 37, 30, 16, 17, 25, 31, 39, 15}]{
        \includegraphics[ width=0.8\columnwidth]{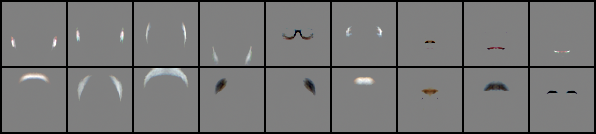}}
\end{center}
\caption{Cosine similarity between localized facial semantics and CelebA facial attributes. }
\vspace{-0.4cm}
\end{figure}